# Decentralized Composite Optimization with Compression

**Yao Li**[*]                                                    LIYAO6@MSU.EDU
*Department of Mathematics*
*Department of Computational Mathematics, Science and Engineering*
*Michigan State University*
*East Lansing, MI 48824, USA*

**Xiaorui Liu**[*]                                               XIAORUI@MSU.EDU
*Department of Computer Science and Engineering*
*Michigan State University*
*East Lansing, MI 48824, USA*

**Jiliang Tang**                                                 TANGJILI@MSU.EDU
*Department of Computer Science and Engineering*
*Michigan State University*
*East Lansing, MI 48824, USA*

**Ming Yan**                                                     MYAN@MSU.EDU
*Department of Computational Mathematics, Science and Engineering*
*Department of Mathematics*
*Michigan State University*
*East Lansing, MI 48824, USA*

**Kun Yuan**                                                     KUN.YUAN@ALIBABA-INC.COM
*DAMO Academy*
*Alibaba (US) Group*
*Bellevue, WA 98004, USA*

## Abstract

Decentralized optimization and communication compression have exhibited their great potential in accelerating distributed machine learning by mitigating the communication bottleneck in practice. While existing decentralized algorithms with communication compression mostly focus on the problems with only smooth components, we study the decentralized stochastic composite optimization problem with a potentially non-smooth component. A <u>P</u>roximal gradient <u>LinEA</u>r convergent <u>D</u>ecentralized algorithm with compression, Prox-LEAD, is proposed with rigorous theoretical analyses in the general stochastic setting and the finite-sum setting. Our theorems indicate that Prox-LEAD works with arbitrary compression precision, and it tremendously reduces the communication cost almost for free. The superiorities of the proposed algorithms are demonstrated through the comparison with state-of-the-art algorithms in terms of convergence complexities and numerical experiments. Our algorithmic framework also generally enlightens the compressed communication on other primal-dual algorithms by reducing the impact of inexact iterations, which might be of independent interest.

**Keywords:** decentralized algorithm, composite optimization, stochastic optimization, communication compression, variance reduction

---

[*] Equal contribution. Authors are listed in alphabetical order.

Yao Li, Xiaorui Liu, Jiliang Tang, Ming Yan, and Kun Yuan

## 1. Introduction

In recent years, the communication cost has become the bottleneck in the distributed training of machine learning models, given that the computation becomes much faster with powerful computing devices such as GPUs and TPUs. Therefore, the communication efficiency gains increasing attention in the algorithm design. On the one hand, decentralized communication has been shown to be an effective and important direction for improving communication efficiency (Lian et al., 2017). On the other hand, various communication compression techniques, such as quantization and sparsification, which are originally developed for centralized settings (Seide et al., 2014; Alistarh et al., 2017; Bernstein et al., 2018; Stich et al., 2018; Karimireddy et al., 2019; Mishchenko et al., 2019; Liu et al., 2020; Tang et al., 2019b), have been shown to be significant in reducing the communication cost for decentralized optimization (Tang et al., 2018a; Reisizadeh et al., 2019a; Koloskova et al., 2019, 2020; Liu et al., 2021; Kovalev et al., 2021a). These works have exhibited the great potential of decentralized optimization with communication compression in speeding up decentralized machine learning.

While existing decentralized algorithms with communication compression mostly focus on the problems with only smooth components, in this paper, we study the decentralized composite optimization problem defined on a connected $n$-node network $\mathcal{G}$ in the form of

$$\underset{\mathbf{x} \in \mathbb{R}^p}{\text{minimize}} \quad \frac{1}{n} \sum_{i=1}^{n} \Big( \underbrace{\mathbb{E}_{\xi_i \sim \mathcal{D}_i} f_i(\mathbf{x}, \xi_i)}_{=: f_i(\mathbf{x})} + r(\mathbf{x}) \Big), \tag{1}$$

where $f_i(\mathbf{x}, \xi_i)$ is the objective function at node $i$ defined on the data $\xi_i$ sampled from the distribution $\mathcal{D}_i$, and $\mathbf{x}$ denotes the model parameters. We use $f_i(\mathbf{x})$ to define the overall local objective function at node $i$, and we will differentiate $f_i(\mathbf{x}, \xi_i)$ and $f_i(\mathbf{x})$ using different inputs. The data distributions $\{\mathcal{D}_i\}$ can be heterogeneous. In other words, the data distributions can be different from node to node, and we do not make the bounded data heterogeneity assumption. The function $f_i(\mathbf{x}, \xi_i)$ is assumed to be convex and smooth, and $r(\mathbf{x})$ is a proper, convex, and possibly non-smooth function that is shared across the nodes. The graph $\mathcal{G}$ encodes the topology of the communication network where information is exchanged along the edges. This problem abstracts many important applications such as regularized empirical risk minimization in statistics and machine learning, and optimal control of multi-agent systems. More specifically, we consider two different settings on the smooth components of the objective function, i.e., the general stochastic setting and the finite-sum setting. In the general stochastic setting, the problem follows (1) where the local functions $\{f_i\}$ are defined as the expectation over the general local sample distributions $\{\mathcal{D}_i\}$. In the finite-sum setting, we consider the discrete distribution on local nodes, and the local functions $\{f_i\}$ are defined as the unweighted average over local samples, $f_i(\mathbf{x}) = \frac{1}{m} \sum_{j=1}^{m} f_{ij}(\mathbf{x})$, where each $f_{ij}$ stands for the loss function defined on the $j$th batch of samples at node $i$. In this work, we assume that the number of batches $m$ is the same for all nodes for simplicity. Note that it can be easily generalized to the case when the nodes have different numbers of batches.

**Contribution.** The main contribution of this paper is a <u>Prox</u>imal gradient <u>LinEA</u>r convergent <u>D</u>ecentralized algorithm with compression, Prox-LEAD, which solves the problem (1). Specifically, the contributions can be summarized as follows:

- We propose the first decentralized stochastic proximal gradient algorithm with compressed communication, Prox-LEAD. It converges linearly up to the neighborhood of the optimal solution in the general stochastic setting. With a diminishing stepsize, the exact solution is achieved at the rate of $\mathcal{O}(1/k)$. In the finite-sum setting, we establish the linear convergence to the exact solution for Prox-LEAD's two variance reduction variants, i.e., Loopless SVRG and SAGA.

- We provide a rigorous theory of Prox-LEAD on the compressed communication and convergence complexities in different settings on the smooth component of the problem. Without the restriction on data heterogeneity and gradient boundedness, Prox-LEAD maintains a comparable convergence rate compared to the uncompressed counterpart. Our theorems indicate that Prox-LEAD works with arbitrary compression precision. Moreover, with reasonably aggressive compression, Prox-LEAD significantly reduces the communication cost almost for free.

- Our algorithmic framework builds bridges between many known algorithms. Without involving compression, it provides stochastic and variance reduction variants of some deterministic algorithms, and it has a better convergence complexity against the existing non-accelerated stochastic decentralized algorithms. When the non-smooth regularizer is absent, it reduces to LEAD and achieves a better convergence rate in the finite-sum setting. The framework also enlightens other primal-dual algorithms to apply compressed communication and reduce the impact of inexact primal and dual iterations.

- We present comprehensive experiments to verify our theorems and the effectiveness of the proposed algorithm in different settings. The comparison with state-of-art algorithms demonstrates the superiority of Prox-LEAD and its stochastic variants. Moreover, it is very robust to parameter tuning, which exhibits great advantages in practice.

The rest of this paper is organized as follows. Section 1.1 and Section 1.2 summarize related works and introduce the notations used in this paper, respectively. In Section 2, the proposed algorithm Prox-LEAD, motivation and derivation, as well as convergence complexities are presented. The assumptions on regularity are introduced in Section 3, and the convergence analyses and major theorems are illustrated in Section 4. Importantly, we also detail the connection with existing algorithms in Section 4.3. Finally, numerical experiments are presented in Section 5.

## 1.1 Related Work

Many algorithms were proposed to solve the decentralized optimization of the average of functions defined over agents in networks. The early decentralized algorithms can be traced back to the work by Tsitsiklis et al. (1986). DGD (Nedic and Ozdaglar, 2009) is the most classical decentralized algorithm. It is intuitive and simple but converges slowly due to the diminishing stepsize that is needed to obtain the optimal solution (Yuan et al., 2016). Its stochastic version D-PSGD (Lian et al., 2017) has been shown effective for training nonconvex deep learning models. Algorithms based on primal-dual formulations or gradient tracking are proposed to eliminate the convergence bias in DGD-type algorithms and improve the convergence rate, such as D-ADMM (Mota et al., 2013), DLM (Ling et al., 2015), EXTRA (Shi et al., 2015a), NIDS (Li et al., 2019), $D^2$ (Tang et al., 2018b), Exact Diffusion (Yuan et al., 2018b), NEXT (Di Lorenzo and Scutari, 2016), DIGing (Nedic et al., 2017), Harnessing (Qu and Li, 2017), SONATA (Scutari and Sun, 2019), GSGT (Pu and Nedić, 2020), OPTRA(Xu et al., 2020a), etc. There are also dual-based methods which apply gradient methods on the dual formulation (Scaman et al., 2017, 2019; Uribe et al., 2020). These algorithms are able to achieve optimal bounds but requires computing the non-trivial gradient of the dual function.

To improve the communication efficiency, communication compression is first applied to decentralized settings by Tang et al. (2018a). It proposes two algorithms, i.e., DCD-SGD and ECD-SGD, which require compression of high accuracy and are not stable with aggressive compression. Reisizadeh et al. (2019a,b) introduce QDGD and QuanTimed-DSGD to achieve exact convergence with small stepsize and the convergence is slow. Deep-Squeeze (Tang et al., 2019a) compensates the compression error to the compression in the next iteration. Motivated by the quantized average consensus algorithms, such as the work in Carli et al. (2010), the quantized gossip algorithm Choco-Gossip (Koloskova et al., 2019) converges linearly to the consensual solution. Combining Choco-Gossip and D-PSGD leads to a decentralized algorithm with compression, Choco-SGD, which converges sublinearly under the strong convexity and gradient boundedness assumptions. Its nonconvex variant is further analyzed in Koloskova et al. (2020). A new compression scheme using the modulo operation is introduced in Lu and De Sa (2020) for decentralized optimization. A general algorithmic framework aiming to maintain the linear convergence of distributed optimization under compressed communication is considered in Magnússon et al. (2020). It requires a contractive property that is not satisfied by many decentralized algorithms including the algorithms proposed in this paper. A preliminary version of this work (Liu et al., 2021) proposes the first linear convergent decentralized algorithm with compression, LEAD. However, the composite problem is not considered in LEAD, and LEAD is deficient in dealing with stochastic gradients. Kovalev et al. (2021a) introduces a linear convergent algorithm for decentralized optimization with communication compression based on a primal-dual decentralized algorithm (Alghunaim and Sayed, 2020). Communication compression is also proposed for gradient tracking algorithms (Li et al., 2021; Xiong et al., 2021; Song et al., 2021), but they require double communication cost since two vectors need to be transmitted in each communication run.

Variance reduction techniques such as SAG (Schmidt et al., 2017), SVRG (Johnson and Zhang, 2013), SAGA (Defazio et al., 2014) and Loopless SVRG (Kovalev et al., 2020) have been introduced to accelerate stochastic optimization problems with finite-sum structure. SVRG-type gradient estimator requires more gradient evaluation but they are memory friendly. SAGA reduces the number of gradient evaluation in each iteration but it requires more memory space. Variance reduction has been applied to decentralized opti-mization (Mokhtari and Ribeiro, 2016; Yuan et al., 2018a; Xin et al., 2020; Li et al., 2020; Hendrikx et al., 2020; Kovalev et al., 2021a).

Decentralized algorithms such as PG-EXTRA (Shi et al., 2015b) and NIDS (Li et al., 2019) are proposed for composite optimization, and the sublinear rate is proved when the smooth component is strongly convex and smooth. Alghunaim et al. (2019) introduces a proximal gradient algorithm (P2D2) and Sun et al. (2019) enhances the convergece rate of SONATA, and the linear convergence rate is proved in both literature when the non-smooth component is shared across all nodes. A proximal unified decentralized algorithm (PUDA) (Alghunaim et al., 2020) and another proximal decentralized algorithmic framework (Xu et al., 2020b) unify many existing algorithms and establish linear convergence for composite optimization when the non-smooth component is shared. A decentralized accelerated proximal gradient descent algorithm in proposed in Ye et al. (2020). However, communication compression and stochastic optimization are not considered in these algorithms.

## 1.2 Notation

We clarify commonly used notation in this section. We use bold lower-case letters to denote column vectors in $\mathbb{R}^p$ and bold upper-case letters for matrices in $\mathbb{R}^{n \times p}$. The lower-case letter with a subscript will be the corresponding row of a matrix denoted by the same letter in the upper-case. For example, in the algorithm, we use $\mathbf{x}_i \in \mathbb{R}^p$ for the local copy of the model parameters at node $i$ and $\mathbf{X} = [\mathbf{x}_1, \mathbf{x}_2, \cdots, \mathbf{x}_n]^\top$ is the matrix, whose rows are these local copies. Throughout the paper, without any specification, we use $\langle \cdot, \cdot \rangle$ as standard vector/matrix inner product, whose form is dependent on context. Similarly, $\| \cdot \|_{\mathbf{P}}$ is a vector/matrix norm defined as $\sqrt{\langle \cdot, \mathbf{P}(\cdot) \rangle}$ for a positive semi-definite matrix $\mathbf{P} \in \mathbb{R}^{n \times n}$, with some specific domain. We use $\mathbf{M}^\dagger$ to denote the pseudo-inverse of a matrix $\mathbf{M} \in \mathbb{R}^{n \times n}$. If $\mathbf{M}$ is symmetric, we use $\lambda_{\max}(\mathbf{M})$, $\lambda_i(\mathbf{M})$ and $\lambda_{\min}(\mathbf{M})$ to denote the largest, the $i$th-largest and the smallest nonzero eigenvalues of $\mathbf{M}$. For the stochastic approximate of the local gradient, we use $\nabla f_i(\mathbf{x}_i, \xi_i) \in \mathbb{R}^p$ as the estimate of the deterministic gradient $\nabla f_i(\mathbf{x}_i) \in \mathbb{R}^p$ of $\mathbf{x}_i$ at node $i$. With the collection of all local estimates $\{\nabla f_i(\mathbf{x}_i, \xi_i)\}_{i=1}^n$ for $\{\nabla f_i(\mathbf{x}_i)\}_{i=1}^n$, we define $\nabla \mathbf{F}(\mathbf{X}) \in \mathbb{R}^{n \times p}$ and $\nabla \mathbf{F}(\mathbf{X}, \xi) \in \mathbb{R}^{n \times p}$ as the compact matrix form of them. The commonly used all-zero vector(matrix) and all-one vector are $\mathbf{0} \in \mathbb{R}^p(\mathbb{R}^{n \times p})$ and $\mathbf{1} \in \mathbb{R}^n$ respectively. The proximal operator with parameter $\eta > 0$ of a function $r : \mathbb{R}^p \to \mathbb{R}$ is $\mathbf{prox}_{\eta r} = \arg\min_{\mathbf{z} \in \mathbb{R}^p} r(\mathbf{z}) + \frac{1}{2\eta} \|\mathbf{z} - \mathbf{x}\|^2$. Finally, we use $[n]$ to replace $\{1, \cdots, n\}$ for abbreviation.

## 2. The Proposed Algorithms

An equivalent form of the problem (1) reformulating the decentralized constraint via a mixing matrix is provided as follows:

$$\mathbf{X}^* = \arg\min_{\mathbf{X} \in \mathbb{R}^{n \times p}} \underbrace{\sum_{i=1}^n f_i(\mathbf{x}_i)}_{=:\mathbf{F}(\mathbf{X})} + \underbrace{\sum_{i=1}^n r(\mathbf{x}_i)}_{=:\mathbf{R}(\mathbf{X})}, \quad \text{s.t. } (\mathbf{I} - \mathbf{W})\mathbf{X} = \mathbf{0}, \tag{2}$$

where $\mathbf{X} = [\mathbf{x}_1, \cdots, \mathbf{x}_n]^\top$ is the collection of local $\mathbf{x}_i$s and $\mathbf{W}$ is a symmetric matrix which restricts the feasible region of the above problem to the subspace $\{\mathbf{1}\mathbf{x}^\top \in \mathbb{R}^{n \times p} \mid \forall \mathbf{x} \in \mathbb{R}^p\}$. The optimal solution $\mathbf{X}^* = \mathbf{1}(\mathbf{x}^*)^\top$ is consensual and provides an optimal solution $\mathbf{x}^* \in \mathbb{R}^p$ to the problem (1). The detailed assumptions on $\mathbf{W}$ are shown below.

**Assumption 1 (Mixing matrix)** *The graph $\mathcal{G} = \{\mathcal{V}, \mathcal{E}\}$ is undirected and connected with $\mathcal{V} = [n] := \{1, 2, \cdots, n\}$. The mixing matrix $\mathbf{W} = [w_{ij}] \in \mathbb{R}^{n \times n}$ is symmetric and satisfies*

  *1. $w_{ij} = 0$ if $i \neq j$ and $(i, j) \notin \mathcal{E}$.*

  *2. $-1 < \lambda_n(\mathbf{W}) \leq \cdots \leq \lambda_2(\mathbf{W}) < \lambda_1(\mathbf{W}) = 1$ and $\mathbf{W}\mathbf{1} = \mathbf{1}$.*

We propose a <u>Prox</u>imal gradient <u>LinEA</u>r convergent <u>D</u>ecentralized algorithm with compressed communication, Prox-LEAD, to solve the problem (2). Our algorithm extends LEAD proposed in Liu et al. (2021) to deal with the non-smooth component via the proximal operator, and to accelerate the stochastic optimization in the finite-sum setting. Moreover, when the compression is absent and the gradient oracle returns full gradient, Prox-LEAD is shown to be covered by the proximal unified decentralized framework in Alghunaim et al. (2020). Algorithm 1 illustrates the prototype of Prox-LEAD in the compact form, and it reduces to LEAD in Algorithm 3 by setting $\mathbf{R} = \mathbf{0}$.

In Algorithm 1, line 6 uses the gradient from the stochastic gradient oracle to create an auxiliary variable $\mathbf{Z}$ as the information to be communicated. The COMM procedure compresses the difference between $\mathbf{Z}$ and a state variable $\mathbf{H}$. The benefit of this difference compression is to reduce the compression error asympototically (Mishchenko et al., 2019; Liu et al., 2021). Since the following variance of the stochastic estimation is dependent on the distance between $\mathbf{Z}^{k+1}$ and $\mathbf{H}^k$,

$$\mathbb{E}\|\hat{\mathbf{Z}}^{k+1} - \mathbf{Z}^{k+1}\|^2 = \mathbb{E}\|\mathcal{Q}(\mathbf{Z}^{k+1} - \mathbf{H}^k) - (\mathbf{Z}^{k+1} - \mathbf{H}^k)\|^2 = \mathcal{O}(\|\mathbf{Z}^{k+1} - \mathbf{H}^k\|),$$

the variance of the compression will vanish as $\mathbf{Z}$ and $\mathbf{H}$ converge to the same point. The updates in line 8 and line 9 essentially compensate the compression error locally. Note that the COMM procedure is first proposed in the preliminary version of this work (Liu et al., 2021), and please to refer to Section 3.1 in Liu et al. (2021) for a detailed explanation for the reduced compression error and implicit error compensation.

Lines 8 and 9 of Algorithm 1 proceed in parallel after the compressed communication is exchanged. The proximal mapping in Line 10 is applied to the rows of $\mathbf{V}^{k+1}$ separately, which is defined as

$$\begin{bmatrix} -\!\!\!-\!\!\!- & (\mathbf{x}_1^{k+1})^\top & -\!\!\!-\!\!\!- \\ & \vdots & \\ -\!\!\!-\!\!\!- & (\mathbf{x}_n^{k+1})^\top & -\!\!\!-\!\!\!- \end{bmatrix} = \begin{bmatrix} -\!\!\!-\!\!\!- & \mathbf{prox}_{\eta r}(\mathbf{v}_1^{k+1})^\top & -\!\!\!-\!\!\!- \\ & \vdots & \\ -\!\!\!-\!\!\!- & \mathbf{prox}_{\eta r}(\mathbf{v}_n^{k+1})^\top & -\!\!\!-\!\!\!- \end{bmatrix}.$$

In the following subsections, we will demonstrate the derivation of Prox-LEAD and how Prox-LEAD inherits the advantages of LEAD intuitively.

---

**Algorithm 1** Prox-LEAD

---

**Input:** Stepsize $\eta$, parameter $\alpha, \gamma$, initial $\mathbf{X}^0, \mathbf{H}^1, \mathbf{D}^1 = \mathbf{0}$

**Output:** $\mathbf{X}^K$ or $1/n \sum_{i=1}^{n} \mathbf{X}_i^K$

   1:   $\mathbf{H}_w^1 = \mathbf{W}\mathbf{H}^1$

   2:   $\mathbf{Z}^1 = \mathbf{X}^0 - \eta\nabla\mathbf{F}(\mathbf{X}^0, \xi^0)$

   3:   $\mathbf{X}^1 = \mathbf{prox}_{\eta\mathbf{R}}(\mathbf{Z}^1)$

   4:   **for** $k = 1, 2, \cdots, K-1$ **do**

   5:      $\mathbf{G}^k = \text{SGO}(\mathbf{X}^k)$

   6:      $\mathbf{Z}^{k+1} = \mathbf{X}^k - \eta\mathbf{G}^k - \eta\mathbf{D}^k$

   7:      $\hat{\mathbf{Z}}^{k+1}, \hat{\mathbf{Z}}_w^{k+1}, \mathbf{H}^{k+1}, \mathbf{H}_w^{k+1} = \mathbf{procedure} \; \text{COMM}(\mathbf{Z}^{k+1}, \mathbf{H}^k, \mathbf{H}_w^k, \alpha)$

   8:      $\mathbf{D}^{k+1} = \mathbf{D}^k + \frac{\gamma}{2\eta}(\hat{\mathbf{Z}}^{k+1} - \hat{\mathbf{Z}}_{\mathbf{W}}^{k+1})$

   9:      $\mathbf{V}^{k+1} = \mathbf{Z}^{k+1} - \frac{\gamma}{2}(\hat{\mathbf{Z}}^{k+1} - \hat{\mathbf{Z}}_{\mathbf{W}}^{k+1})$

 10:      $\mathbf{X}^{k+1} = \mathbf{prox}_{\eta\mathbf{R}}\left(\mathbf{V}^{k+1}\right)$

 11:   **end for**

---

---

Compressed Communication Procedure (COMM)

---

   1:   **procedure** $\text{COMM}(\mathbf{Z}^{k+1}, \mathbf{H}^k, \mathbf{H}_w^k, \alpha)$

   2:      $\mathbf{Q}^k = \mathcal{Q}(\mathbf{Z}^{k+1} - \mathbf{H}^k)$          ▷ Compression

   3:      $\hat{\mathbf{Z}}^{k+1} = \mathbf{H}^k + \mathbf{Q}^k$

   4:      $\hat{\mathbf{Z}}_w^{k+1} = \mathbf{H}_w^k + \mathbf{W}\mathbf{Q}^k$        ▷ Communication

   5:      $\mathbf{H}^{k+1} = (1-\alpha)\mathbf{H}^k + \alpha\hat{\mathbf{Z}}$

   6:      $\mathbf{H}_w^{k+1} = (1-\alpha)\mathbf{H}_w^k + \alpha\hat{\mathbf{Z}}_w$

   7:      **Return:** $\hat{\mathbf{Z}}^{k+1}, \hat{\mathbf{Z}}_w^{k+1}, \mathbf{H}^{k+1}, \mathbf{H}_w^{k+1}$

   8:   **end procedure**

---

## 2.1 LEAD as Inexact PDHG

We firstly provide a new view of LEAD from the perspective of operator splitting. Consider the consensus optimization problem

$$\min_{\mathbf{X}\in\mathbb{R}^{n\times p}} \mathbf{F}(\mathbf{X}) \quad \text{s.t. } \mathbf{B}^{\frac{1}{2}}\mathbf{X} = \mathbf{0},$$

and the corresponding saddle-point formulation

$$\min_{\mathbf{X}\in\mathbb{R}^{n\times p}} \max_{\mathbf{S}\in\mathbb{R}^{n\times p}} \mathbf{F}(\mathbf{X}) + \langle\mathbf{B}^{\frac{1}{2}}\mathbf{X}, \mathbf{S}\rangle \tag{3}$$

where $\mathbf{S}$ is the Lagrangian multiplier and $\mathbf{B} = \frac{\mathbf{I}-\mathbf{W}}{2}$. Many operator splitting schemes can be applied to this problem such as Douglas-Rachford scheme in Eckstein and Bertsekas (1992) and Chambolle-Pock in Chambolle and Pock (2011). We are using PDHG (Primal

Dual Hybrid Gradient) proposed by Zhu and Chan (2008) and get

$$
\begin{cases}
\mathbf{X}^{k+1} = \underset{\mathbf{X} \in \mathbb{R}^{n \times p}}{\arg\min} \ \mathbf{F}(\mathbf{X}) + \langle \mathbf{B}^{\frac{1}{2}}\mathbf{X}, \mathbf{S}^{k} \rangle, \\
\mathbf{S}^{k+1} = \mathbf{S}^{k} + \lambda \mathbf{B}^{\frac{1}{2}}\mathbf{X}^{k+1}.
\end{cases}
$$

The iteration first solves the minimization problem with fixed dual variable $\mathbf{S}^{k}$, and then updates the dual step using gradient ascent with stepsize $\lambda$. This iteration can also be derived from AMA (Alternating Minimization Algorithm) proposed in Tseng (1991) so the convergence is guaranteed if $\mathbf{F}$ is strongly convex. Note that the $\mathbf{X}$-subproblem is non-trivial to solve. Therefore, we solve it inexactly by two step gradient descent updates with stepsize $\eta$. By algebraic reformulation, we obtain

$$
\begin{cases}
\mathbf{X}^{k+1} = \mathbf{X}^{k} - \eta \nabla \mathbf{F}(\mathbf{X}^{k}) - \eta \mathbf{B}^{\frac{1}{2}}\mathbf{S}^{k}, \\
\overline{\mathbf{X}}^{k+1} = \mathbf{X}^{k+1} - \eta \nabla \mathbf{F}(\mathbf{X}^{k+1}) - \eta \mathbf{B}^{\frac{1}{2}}\mathbf{S}^{k}, \\
\mathbf{S}^{k+1} = \mathbf{S}^{k} + \lambda \mathbf{B}^{\frac{1}{2}}\overline{\mathbf{X}}^{k+1}.
\end{cases}
$$

To exploit the fact that the gradient computation $\nabla \mathbf{F}(\mathbf{X}^{k})$ can be shared between iterations, by switching the order, adjusting the index notations and letting $\mathbf{D} = \mathbf{B}^{\frac{1}{2}}\mathbf{S}$, we obtain the following iteration

$$
\begin{cases}
\overline{\mathbf{X}}^{k+1} = \mathbf{X}^{k} - \eta \nabla \mathbf{F}(\mathbf{X}^{k}) - \eta \mathbf{D}^{k}, \\
\mathbf{D}^{k+1} = \mathbf{D}^{k} + \dfrac{\lambda}{2}(\mathbf{I} - \mathbf{W})\overline{\mathbf{X}}^{k+1}, \\
\mathbf{X}^{k+1} = \mathbf{X}^{k} - \eta \nabla \mathbf{F}(\mathbf{X}^{k}) - \eta \mathbf{D}^{k+1}.
\end{cases}
\tag{4}
$$

It can be shown that the iteration (4) is equivalent to LEAD when there is no compression and the full gradient is used. Moreover, it is exactly the iteration of PAPC in Loris and Verhoeven (2011); Chen et al. (2013) applied to the problem (3).

---

**Algorithm 3** LEAD

---

**Input:** Stepsize $\eta$, parameter $\alpha, \gamma$, initial $\mathbf{X}^{0}, \mathbf{H}^{1}, \mathbf{D}^{1} = \mathbf{0}$

**Output:** $\mathbf{X}^{K}$ or $1/n \sum_{i=1}^{n} \mathbf{X}_{i}^{K}$

1: $\mathbf{H}_{w}^{1} = \mathbf{W}\mathbf{H}^{1}$
2: $\mathbf{X}^{1} = \mathbf{X}^{0} - \eta \nabla \mathbf{F}(\mathbf{X}^{0}, \xi^{0})$
3: **for** $k = 1, 2, \cdots, K-1$ **do**
4:      $\mathbf{Z}^{k+1} = \mathbf{X}^{k} - \eta \nabla \mathbf{F}(\mathbf{X}^{k}, \xi^{k}) - \eta \mathbf{D}^{k}$
5:      $\hat{\mathbf{Z}}^{k+1}, \hat{\mathbf{Z}}_{w}^{k+1}, \mathbf{H}^{k+1}, \mathbf{H}_{w}^{k+1} = $ **procedure** COMM$(\mathbf{Z}^{k+1}, \mathbf{H}^{k}, \mathbf{H}_{w}^{k}, \alpha)$
6:      $\mathbf{D}^{k+1} = \mathbf{D}^{k} + \frac{\gamma}{2\eta}(\hat{\mathbf{Z}}^{k+1} - \hat{\mathbf{Z}}_{\mathbf{W}}^{k+1})$
7:      $\mathbf{X}^{k+1} = \mathbf{X}^{k} - \eta \nabla \mathbf{F}(\mathbf{X}^{k}, \xi^{k}) - \eta \mathbf{D}^{k+1}$
8: **end for**

---

By compressing the only communication step $\mathbf{D}$ via the procedure COMM, we obtain LEAD as showed in Algorithm 3. We regard LEAD as inexact PDHG in terms of the inexact primal and dual updates. The compression error can be controlled and converges to 0 given the bounded noise-to-signal ratio assumption on the stochastic compression operator.

**Assumption 2 (Unbiased compression operator)** *The stochastic operator $\mathcal{Q} : \mathbb{R}^p \to \mathbb{R}^p$ is unbiased, i.e., $\mathbb{E}\mathcal{Q}(\mathbf{x}) = \mathbf{x}$, and there exists $C \geq 0$ such that*

$$\mathbb{E}\|\mathbf{x} - \mathcal{Q}(\mathbf{x})\|^2 \leq C\|\mathbf{x}\|^2, \ \forall \mathbf{x} \in \mathbb{R}^p.$$

*In particular, when $C = 0$, we treat $\mathcal{Q}$ as the identity operator.*

The following theorem from Liu et al. (2021) provides the theoretical guarantee on the iteration.

**Theorem 1** *Let $\{\mathbf{X}^k\}$ be the sequence generated from iteration (4) with compression operator satisfying Assumption 2. Assume $\mathbf{F}$ is $\mu$-strongly convex and $L$-smooth, by taking $\eta = \frac{1}{L}$, $\alpha = \mathcal{O}\left(\frac{1}{(1+C)\kappa_f}\right)$ and $\lambda = \min\left\{\frac{\mu}{C\lambda_{\max}(\mathbf{I}-\mathbf{W})}, \frac{L}{(1+3C)\lambda_{\max}(\mathbf{I}-\mathbf{W})}\right\}$, then $\{\mathbf{X}^k\}$ converges to the $\epsilon$-accurate solution with the iteration complexity[1]*

$$\widetilde{\mathcal{O}}\left((1+C)(\kappa_f + \kappa_g) + C\kappa_f\kappa_g\right),$$

*where $\kappa_f = \frac{L}{\mu}$ and $\kappa_g = \frac{\lambda_{\max}(\mathbf{I}-\mathbf{W})}{\lambda_{\min}(\mathbf{I}-\mathbf{W})}$.*

## 2.2 Prox-LEAD as Inexact PUDA

Returning to the problems with the non-smooth regularizer, the problem (3) becomes the following three operators splitting

$$\underset{\mathbf{X} \in \mathbb{R}^{n \times p}}{\text{minimize}} \ \mathbf{F}(\mathbf{X}) + \iota_{\mathbf{0}}(\mathbf{B}^{\frac{1}{2}}\mathbf{X}) + \mathbf{R}(\mathbf{X}). \tag{5}$$

Many existing schemes such as Condat-Vu in Condat (2013); Vu (2013), PDFP in Chen et al. (2016) and PD3O in Yan (2018) are applicable to this problem but we choose to adapt the inexact PDHG with a single proximal gradient step and derive the following iteration which is not the realization of any scheme mentioned above.

$$\left|\begin{array}{l} \overline{\mathbf{X}}^{k+1} = \mathbf{X}^k - \eta\nabla\mathbf{F}(\mathbf{X}^k) - \eta\mathbf{D}^k, \\[2mm] \mathbf{D}^{k+1} = \mathbf{D}^k + \dfrac{\lambda}{2}(\mathbf{I} - \mathbf{W})\overline{\mathbf{X}}^{k+1}, \\[2mm] \mathbf{V}^{k+1} = \mathbf{X}^k - \eta\nabla\mathbf{F}(\mathbf{X}^k) - \eta\mathbf{D}^{k+1} = \left(\mathbf{I} - \dfrac{\eta\lambda}{2}(\mathbf{I} - \mathbf{W})\right)\overline{\mathbf{X}}^{k+1}, \\[2mm] \mathbf{X}^{k+1} = \mathbf{prox}_{\eta\mathbf{R}}(\mathbf{V}^{k+1}). \end{array}\right. \tag{6}$$

Compared to LEAD in the form of (4), the third step is reformulated to fully depend on $\overline{\mathbf{X}}$ but the number of communication is unchanged, and a proximal map is applied on $\mathbf{V}$ directly in the final step. The iteration (6) can be shown as a special case of PUDA in Alghunaim et al. (2020) which has global linear convergence for strongly convex $\mathbf{F}$.

---

[1] $\widetilde{\mathcal{O}}$-notation hides polylogarithmic factors.

The benefit of this iteration is to maintain the consensus of $\overline{\mathbf{X}}$ in optimality, which further implies the consensus of $\mathbf{V}$ and $\mathbf{X}$. As shown in Section 4, the consensus of $\overline{\mathbf{X}}$ is the key to the linear convergence of Prox-LEAD. It also explains why we need the same non-smooth function $r$ for all nodes.

Similar to LEAD, if we compress the only communication step involving $\overline{\mathbf{X}}$ via COMM procedure, $(\mathbf{D}, \mathbf{V})$ is updated by the inexact information with the controllable compression error. Therefore, we regard Prox-LEAD as inexact PUDA in terms of the inexact dual and proximal steps.

### 2.3 Stochastic Gradient Oracle

As shown in Algorithm 1, Prox-LEAD uses a stochastic gradient oracle (SGO) to estimate the gradient, and different stochastic estimators are listed in Table 1 for gradient estimation.

| | The finite-sum setting | |
| The general setting | Loopless SVRG | SAGA |
| --- | --- | --- |
| Sample $\xi_i \sim \mathcal{D}_i$ | Sample $l \in [m] \sim \mathcal{P}_i$ randomly | |
| $\mathbf{g}_i = \nabla f_i(\mathbf{x}_i, \xi_i).$ | Sample $\omega \in \{0,1\} \sim Bernoulli(p),$ $\mathbf{g}_i = \frac{1}{mp_{il}}(\nabla f_{il}(\mathbf{x}_i) - \nabla f_{il}(\tilde{\mathbf{x}}_i))$ $\quad + \nabla f_i(\tilde{\mathbf{x}}_i),$ $\tilde{\mathbf{x}}_i = \omega \cdot \mathbf{x}_i + (1-\omega) \cdot \tilde{\mathbf{x}}_i.$ | $\mathbf{g}_i = \frac{1}{mp_{il}}(\nabla f_{il}(\mathbf{x}_i) - \nabla f_{il}(\tilde{\mathbf{x}}_{il}))$ $\quad + \frac{1}{m}\sum_{j=1}^{m} \nabla f_{ij}(\tilde{\mathbf{x}}_{ij}),$ $\tilde{\mathbf{x}}_{il} = \mathbf{x}_i.$ |

Table 1: Stochastic gradient oracle (SGO)

The stochastic gradient oracle returns three types of estimation. In the general setting, each node uses sample distribution $\mathcal{D}_i$ to provide an unbiased stochastic gradient and the variance exists. In the finite-sum setting, we assume each node will construct a discrete distribution $\mathcal{P}_i = \{p_{il} : l \in [m]\}$ for $m$ mini-batches and the stochastic gradient will be corrected by two different variance reduction schemes: Loopless SVRG and SAGA.

For Loopless SVRG (LSVRG), each node will have a reference point $\tilde{x}_i$ where the full gradient is evaluated after a random period. A random variable $l \in [m]$ will be sampled first with distribution $\mathcal{P}_i$, then a Bernoulli random variable $\omega$ will be sampled to determine whether the reference point will be update or not. If $\omega = 1$, the reference is replaced by the latest $\mathbf{x}_i$, otherwise unchanged. For SAGA, each node will have $m$ reference points, $\{\tilde{x}_{ij} : j \in [m]\}$. After the index $l$ is sampled, $\tilde{\mathbf{x}}_{il}$ will be replaced by the latest $\mathbf{x}_i$ while remaining reference points are unchanged. Both schemes use reference points to correct the gradient by variance reduction. Loopless SVRG is memory-friendly but requires more gradient evaluations. Empirically, SAGA converges faster in terms of the number of gradient evaluations, but it requires more memory space. The following Table 2 summarizes the convergence complexity of Prox-LEAD in different settings.

| Algorithm | $\alpha, \eta, \gamma$ | Convergence complexity |
|:---:|:---:|:---:|
| Prox-LEAD Theorem 5 | fixed | $\widetilde{\mathcal{O}}((1+C)(\kappa_f + \kappa_g) + \sqrt{C}(1+C)\kappa_f\kappa_g)$ |
| Prox-LEAD Theorem 7 | $\mathcal{O}(\frac{1}{k})$ | $\mathcal{O}\left(\left((1+C)^2\kappa_f\kappa_g + \frac{\sigma^2}{L^2}(1+C)^4\kappa_f^2\kappa_g^2\right)\frac{1}{\epsilon}\right)$ |
| Prox-LEAD LSVRG Theorem 8 | fixed | $\widetilde{\mathcal{O}}((1+C)(\kappa_f + \kappa_g) + \sqrt{C}(1+C)\kappa_f\kappa_g + p^{-1})$ |
| Prox-LEAD SAGA Theorem 9 | fixed | $\widetilde{\mathcal{O}}((1+C)(\kappa_f + \kappa_g) + \sqrt{C}(1+C)\kappa_f\kappa_g + m)$ |

Table 2: Summary of the convergence compleixty for Prox-LEAD to achieve $\epsilon$-accuracy. The first row is the complexity with the full gradient.

When the compression procedure and the regularizer are removed, the complexities of LEAD LSVRG and LEAD SAGA are reduced to $\widetilde{\mathcal{O}}(\kappa_f + \kappa_g + p^{-1})$ and $\widetilde{\mathcal{O}}(\kappa_f + \kappa_g + m)$ respectively, which are better than LessBit-Option D in Kovalev et al. (2021a) in terms of the conditional numbers, and it improves over the stochastic LEAD in Liu et al. (2021).

## 3. Assumptions on Regularity

In the general stochastic setting, each $f_i(\mathbf{x})$ in problem (1) is the expectation of the local loss function under the sample distribution $\mathcal{D}_i$, and we make the following inter-node assumption.

**Assumption 3 (Locally bounded gradient variance)** *In the general stochastic setting, each local stochastic gradient $\nabla f_i(\mathbf{x}, \xi_i)$ is an unbiased estimate, i.e., $\mathbb{E}\nabla f_i(\mathbf{x}, \xi_i) = \nabla f_i(x)$, and satisfies*

$$\mathbb{E}\|\nabla f_i(\mathbf{x}^*, \xi_i) - \nabla f_i(\mathbf{x}^*)\|^2 \leq \sigma_i^2,$$

*where $\mathbf{x}^*$ is the optimal solution to the problem (1).*

This locally bounded variance at the optimal point is strictly weaker than the uniformly bounded variance assumption made in Liu et al. (2021).

Given a smooth convex function $f$, we define the Bregman distance with respect to $f$ as

$$\mathrm{V}_f(\mathbf{x}, \mathbf{y}) = f(\mathbf{x}) - f(\mathbf{y}) - \langle \nabla f(\mathbf{y}), \mathbf{x} - \mathbf{y} \rangle, \quad \forall \mathbf{x}, \mathbf{y} \in \mathbb{R}^p.$$

With the above definition, we impose different assumptions to the regularity of smooth function component in two settings.

**Assumption 4 (Strong convexity and smoothness)** *Each $f_i$ is a smooth, $\mu$-strongly convex function, i.e., $\forall \mathbf{x}, \mathbf{y} \in \mathbb{R}^p$,*

$$\mathrm{V}_{f_i}(\mathbf{x}, \mathbf{y}) = f_i(\mathbf{x}) - f_i(\mathbf{y}) - \langle \nabla f_i(\mathbf{y}), \mathbf{x} - \mathbf{y} \rangle \geq \frac{\mu}{2}\|\mathbf{x} - \mathbf{y}\|^2. \tag{7}$$

*In the general stochastic setting, each $f_i$ is $L$-smooth in expectation, i.e., $\forall \mathbf{x}, \mathbf{y} \in \mathbb{R}^p$,*

$$\mathbb{E}\|\nabla f_i(\mathbf{x}, \xi_i) - \nabla f_i(\mathbf{y}, \xi_i)\|^2 \leq 2L[f_i(\mathbf{x}) - f_i(\mathbf{y}) - \langle \nabla f_i(\mathbf{y}), \mathbf{x} - \mathbf{y} \rangle] = 2L\mathrm{V}_{f_i}(\mathbf{x}, \mathbf{y}). \tag{8}$$

In the finite-sum setting, the $L$-smoothness is imposed on each $f_{ij}$ instead, i.e., $\forall \mathbf{x}, \mathbf{y} \in \mathbb{R}^p$,

$$\|\nabla f_{ij}(\mathbf{x}) - \nabla f_{ij}(\mathbf{y})\|^2 \leq 2L[f_{ij}(\mathbf{x}) - f_{ij}(\mathbf{y}) - \langle \nabla f_{ij}(\mathbf{y}), \mathbf{x} - \mathbf{y} \rangle] = 2L \mathrm{V}_{f_{ij}}(\mathbf{x}, \mathbf{y}). \quad (9)$$

Note the inequalities in (7) and (8) are equivalent to, $\forall \mathbf{X}, \mathbf{Y} \in \mathbb{R}^{n \times p}$,

$$\mathrm{V}_{\mathbf{F}}(\mathbf{X}, \mathbf{Y}) = \mathbf{F}(\mathbf{X}) - \mathbf{F}(\mathbf{Y}) - \langle \nabla \mathbf{F}(\mathbf{Y}), \mathbf{X} - \mathbf{Y} \rangle \geq \frac{\mu}{2} \|\mathbf{X} - \mathbf{Y}\|^2,$$

$$\mathbb{E}\|\nabla \mathbf{F}(\mathbf{X}, \xi) - \nabla \mathbf{F}(\mathbf{Y}, \xi)\|^2 \leq 2L \sum_{i=1}^{n} \mathrm{V}_{f_i}(\mathbf{x}_i, \mathbf{y}_i) = 2L \mathrm{V}_{\mathbf{F}}(\mathbf{X}, \mathbf{Y}).$$

**Remark 2** *In Gower et al. (2019), the above two types of $L$-smoothness assumptions are discussed, and the latter is used to derive an improved convergence rate of SGD. The expected $L$-smoothness is shown to be significantly weaker than the most commonly-used $L$-smoothness of $f_i(\mathbf{x}, \xi_i)$ for all $\xi_i$.*

## 4. Convergence Analysis

In this section, we present the convergence of Prox-LEAD in both stochastic scenarios. We first show two fundamental lemmas regarding the conditional expectation on the compression operator, then we present the two scenarios in Sections 4.1 and 4.2, respectively. More specifically, in Section 4.1, the linear convergence to the neighborhood will be shown under the general stochastic setting, while in Section 4.2, two variance reduction schemes are used to exploit the exact linear convergence in the problems with finite-sum structure.

The stochastic actions such as compression and gradient estimation generate two sequences of $\sigma$-algebra where the stochastic variables in this procedure are adapted. We use $\mathcal{F}^k$ to denote the $\sigma$-algebra of gradient estimation at $k$th step and $\mathcal{H}^k$ is the $\sigma$-algebra of stochastic compression at the same step. $\{\mathcal{F}^k\}$ and $\{\mathcal{H}^k\}$ satisfy

$$\mathcal{F}^1 \subset \mathcal{H}^1 \subset \mathcal{F}^2 \subset \mathcal{H}^2 \subset \cdots \subset \mathcal{F}^k \subset \mathcal{H}^k \subset \cdots.$$

With these notations, we can clarify the stochastic dependencies among the variables generated by the algorithm. For example, tuple $(\mathbf{G}^k, \mathbf{Z}^{k+1})$ is measurable in $\mathcal{F}^k$ and tuple $(\hat{\mathbf{Z}}^{k+1}, \mathbf{H}^{k+1}, \mathbf{D}^{k+1}, \mathbf{V}^{k+1}, \mathbf{X}^{k+1})$ is measurable in $\mathcal{H}^k$.

Throughout the section, we use $\mathbb{E}$ to denote the conditional expectations $\mathbb{E}_{\mathcal{F}^k}$ and $\mathbb{E}_{\mathcal{H}^k}$ given the context for simplicity. Then we define some auxiliary constants related to the optimal solution $\mathbf{X}^*$ to the problem (2). $\mathbf{X}^*$ is consensual, i.e., each row $\mathbf{x}^*$ solves the problem (1). We let $\mathbf{z}^* = \mathbf{x}^* - (\eta/n) \sum_{i=1}^{n} \nabla f_i(\mathbf{x}^*)$, which is in the pre-image of $\mathbf{prox}_{\eta r}$ at $\mathbf{x}^*$. In addition, we let

$$\mathbf{Z}^* = \mathbf{1}(\mathbf{z}^*)^\top = \mathbf{X}^* - \frac{\eta}{n} \mathbf{1}\mathbf{1}^\top \nabla \mathbf{F}(\mathbf{X}^*), \quad (10)$$

$$\mathbf{D}^* = \frac{1}{\eta}(\mathbf{Z}^* - \mathbf{X}^*) + \nabla \mathbf{F}(\mathbf{X}^*) = \left(\mathbf{I} - \frac{1}{n}\mathbf{1}\mathbf{1}^\top\right) \nabla \mathbf{F}(\mathbf{X}^*). \quad (11)$$

To show the convergence of the proposed algorithm, we characterizes the decrease of $\|\mathbf{Z}^k - \mathbf{Z}^*\|$, $\|\mathbf{D}^k - \mathbf{D}^*\|_{(\mathbf{I}-\mathbf{W})^\dagger}$, $\|\mathbf{X}^k - \mathbf{X}^*\|$, and $\|\mathbf{H}^k - \mathbf{Z}^*\|$. The convergence of $\mathbf{H}^k$ to $\mathbf{Z}^*$ shows the decrease of the compression error to zero. The following lemma shows the connection of those values for the $k$ and $k+1$-th iteration.

**Lemma 3 (One-step progress)** *Let $\{(\mathbf{Z}^k, \mathbf{D}^k, \mathbf{X}^k)\}$ be the sequence generated from Algorithm 1. Under Assumption 1, taking the expectation conditioned on the stochastic compression operator at the $k$-th iteration, we have for any $\gamma \leq \frac{2}{\lambda_{\max}(\mathbf{I}-\mathbf{W})}$,*

$$
\|\mathbf{Z}^{k+1} - \mathbf{Z}^*\|^2 = \|\mathbf{X}^k - \mathbf{X}^* - \eta\mathbf{G}^k + \eta\nabla\mathbf{F}(\mathbf{X}^*)\|^2 + \eta^2\|\mathbf{D}^k - \mathbf{D}^*\|^2
$$
$$
- 2\eta\langle\mathbf{D}^k - \mathbf{D}^*, \mathbf{X}^k - \mathbf{X}^* - \eta\mathbf{G}^k + \eta\nabla\mathbf{F}(\mathbf{X}^*)\rangle, \tag{12}
$$

$$
\mathbb{E}\|\mathbf{D}^{k+1} - \mathbf{D}^*\|^2_{(\mathbf{I}-\mathbf{W})^\dagger} = \|\mathbf{D}^k - \mathbf{D}^*\|^2_{(\mathbf{I}-\mathbf{W})^\dagger} - \gamma\|\mathbf{D}^k - \mathbf{D}^*\|^2
$$
$$
+ \frac{\gamma^2}{4\eta^2}\|\mathbf{Z}^{k+1} - \mathbf{Z}^*\|^2_{\mathbf{I}-\mathbf{W}} + \frac{\gamma^2}{4\eta^2}\mathbb{E}\|\hat{\mathbf{Z}}^{k+1} - \mathbf{Z}^{k+1}\|^2_{\mathbf{I}-\mathbf{W}}
$$
$$
+ \frac{\gamma}{\eta}\langle\mathbf{D}^k - \mathbf{D}^*, \mathbf{X}^k - \mathbf{X}^* - \eta\mathbf{G}^k + \eta\nabla\mathbf{F}(\mathbf{X}^*)\rangle, \tag{13}
$$

$$
\mathbb{E}\|\mathbf{X}^{k+1} - \mathbf{X}^*\|^2 \leq \|\mathbf{Z}^{k+1} - \mathbf{Z}^*\|^2_{\mathbf{I}-\frac{\gamma}{2}(\mathbf{I}-\mathbf{W})} + \frac{\gamma^2}{4}\mathbb{E}\|\hat{\mathbf{Z}}^{k+1} - \mathbf{Z}^{k+1}\|^2_{(\mathbf{I}-\mathbf{W})^2}. \tag{14}
$$

**Proof** (i) The equality (12) is shown from Line 6 of Algorithm 1 and (11) directly.

(ii) Note that $(\mathbf{I} - \mathbf{W})\mathbf{Z}^* = \mathbf{0}$. Then from Line 8 of Algorithm 1, we have

$$
\mathbf{D}^{k+1} - \mathbf{D}^* = \mathbf{D}^k - \mathbf{D}^* + \frac{\gamma}{2\eta}(\mathbf{I} - \mathbf{W})(\hat{\mathbf{Z}}^{k+1} - \mathbf{Z}^*),
$$

which gives

$$
\mathbb{E}\|\mathbf{D}^{k+1} - \mathbf{D}^*\|^2_{(\mathbf{I}-\mathbf{W})^\dagger}
$$
$$
= \|\mathbf{D}^k - \mathbf{D}^*\|^2_{(\mathbf{I}-\mathbf{W})^\dagger} + \frac{\gamma}{\eta}\langle\mathbf{D}^k - \mathbf{D}^*, \mathbf{Z}^{k+1} - \mathbf{Z}^*\rangle + \frac{\gamma^2}{4\eta^2}\|\mathbf{Z}^{k+1} - \mathbf{Z}^*\|^2_{\mathbf{I}-\mathbf{W}}
$$
$$
+ \frac{\gamma^2}{4\eta^2}\mathbb{E}\|\hat{\mathbf{Z}}^{k+1} - \mathbf{Z}^{k+1}\|^2_{\mathbf{I}-\mathbf{W}}
$$
$$
= \|\mathbf{D}^k - \mathbf{D}^*\|^2_{(\mathbf{I}-\mathbf{W})^\dagger} + \frac{\gamma}{\eta}\langle\mathbf{D}^k - \mathbf{D}^*, \mathbf{X}^k - \mathbf{X}^* - \eta\mathbf{G}^k + \eta\nabla\mathbf{F}(\mathbf{X}^*)\rangle
$$
$$
- \gamma\|\mathbf{D}^k - \mathbf{D}^*\|^2 + \frac{\gamma^2}{4\eta^2}\|\mathbf{Z}^{k+1} - \mathbf{Z}^*\|^2_{\mathbf{I}-\mathbf{W}} + \frac{\gamma^2}{4\eta^2}\mathbb{E}\|\hat{\mathbf{Z}}^{k+1} - \mathbf{Z}^{k+1}\|^2_{\mathbf{I}-\mathbf{W}},
$$

where the second equality comes from Line 6 of Algorithm 1.

(iii) From the definition of $\mathbf{Z}^*$, we have

$$
\mathbf{X}^* = \mathbf{prox}_{\eta\mathbf{R}}(\mathbf{Z}^*).
$$

Therefore

$$
\mathbb{E}\|\mathbf{X}^{k+1} - \mathbf{X}^*\|^2 = \|\mathbf{prox}_{\eta\mathbf{R}}(\mathbf{V}^{k+1}) - \mathbf{prox}_{\eta\mathbf{R}}(\mathbf{Z}^*)\|^2
$$
$$
\leq \mathbb{E}\|\mathbf{V}^{k+1} - \mathbf{Z}^*\|^2
$$
$$
= \mathbb{E}\left\|\mathbf{Z}^{k+1} - \mathbf{Z}^* - \left(\frac{\gamma}{2}(\mathbf{I} - \mathbf{W})(\hat{\mathbf{Z}}^{k+1} - \mathbf{Z}^*)\right)\right\|^2
$$
$$
= \left\|\left(\mathbf{I} - \frac{\gamma}{2}(\mathbf{I} - \mathbf{W})\right)(\mathbf{Z}^{k+1} - \mathbf{Z}^*)\right\|^2 + \frac{\gamma^2}{4}\mathbb{E}\|(\mathbf{I} - \mathbf{W})\hat{\mathbf{Z}}^{k+1} - \mathbf{Z}^{k+1}\|^2
$$
$$
\leq \|\mathbf{Z}^{k+1} - \mathbf{Z}^*\|^2_{\mathbf{I}-\frac{\gamma}{2}(\mathbf{I}-\mathbf{W})} + \frac{\gamma^2}{4}\mathbb{E}\|(\mathbf{I} - \mathbf{W})\hat{\mathbf{Z}}^{k+1} - \mathbf{Z}^{k+1}\|^2,
$$

where the first inequality comes from the non-expansiveness of $\mathbf{prox}_{\eta\mathbf{R}}$ and the last inequality follows from

$$
\begin{aligned}
\left(\mathbf{I} - \frac{\gamma}{2}(\mathbf{I} - \mathbf{W})\right)^2 &= \mathbf{I} - \gamma(\mathbf{I} - \mathbf{W}) + \frac{\gamma^2}{4}(\mathbf{I} - \mathbf{W})^2 \\
&= \mathbf{I} - \frac{\gamma}{2}(\mathbf{I} - \mathbf{W}) + \frac{\gamma}{2}(\mathbf{I} - \mathbf{W})^{\frac{1}{2}}\left(-\mathbf{I} + \frac{\gamma}{2}(\mathbf{I} - \mathbf{W})\right)(\mathbf{I} - \mathbf{W})^{\frac{1}{2}} \\
&\preccurlyeq \mathbf{I} - \frac{\gamma}{2}(\mathbf{I} - \mathbf{W}),
\end{aligned}
$$

since $\frac{\gamma}{2}(\mathbf{I} - \mathbf{W}) \preccurlyeq \mathbf{I}$. The inequality (14) is obtained. ∎

The next lemma builds the critical inequality for Algorithm 1 and serves as the key step in the proofs.

**Lemma 4 (The key inequality)** *Under Assumptions 1 and 2, we choose* $\alpha \in (0, (1 + C)^{-1})$ *and*

$$
\gamma \in \left(0, \frac{1}{\lambda_{\max}(\mathbf{I} - \mathbf{W})}\min\left\{\frac{\Delta(\alpha)}{\sqrt{C}}, 2(1 - \sqrt{C}\alpha)\right\}\right), \quad \Delta(\alpha) := \alpha - (1 + C)\alpha^2 > 0.
$$

*Then the sequence* $\{(\mathbf{X}^k, \mathbf{D}^k, \mathbf{H}^k)\}$ *generated by Algorithm 1 satisfies*

$$
\begin{aligned}
&M\mathbb{E}\|\mathbf{X}^{k+1} - \mathbf{X}^*\|^2 + \frac{2\eta^2}{\gamma}\mathbb{E}\|\mathbf{D}^{k+1} - \mathbf{D}^*\|^2_{(\mathbf{I} - \mathbf{W})^\dagger} + \sqrt{C}\mathbb{E}\|\mathbf{H}^{k+1} - \mathbf{Z}^*\|^2 \\
&\leq \frac{2\eta^2}{\gamma}\left(1 - \frac{\gamma}{2}\lambda_{\min}(\mathbf{I} - \mathbf{W})\right)\|\mathbf{D}^k - \mathbf{D}^*\|^2_{(\mathbf{I} - \mathbf{W})^\dagger} + \|\mathbf{X}^k - \mathbf{X}^* - \eta\mathbf{G}^k + \eta\nabla\mathbf{F}(\mathbf{X}^*)\|^2 \\
&\quad + \sqrt{C}(1 - \alpha)\|\mathbf{H}^k - \mathbf{Z}^*\|^2.
\end{aligned} \tag{15}
$$

*where the expectation is conditioned on the stochastic compression at* $k$-*th step and*

$$
M := 1 - \frac{\sqrt{C}\alpha}{1 - \frac{\gamma}{2}\lambda_{\max}(\mathbf{I} - \mathbf{W})}.
$$

**Proof** Letting (12) + (13)$\times\frac{2\eta^2}{\gamma}$, we get

$$
\begin{aligned}
&\|\mathbf{Z}^{k+1} - \mathbf{Z}^*\|^2_{\mathbf{I} - \frac{\gamma}{2}(\mathbf{I} - \mathbf{W})} + \frac{2\eta^2}{\gamma}\mathbb{E}\|\mathbf{D}^{k+1} - \mathbf{D}^*\|^2_{(\mathbf{I} - \mathbf{W})^\dagger} \\
&= \|\mathbf{X}^k - \mathbf{X}^* - \eta\mathbf{G}^k + \eta\nabla\mathbf{F}(\mathbf{X}^*)\|^2 + \frac{2\eta^2}{\gamma}\|\mathbf{D}^k - \mathbf{D}^*\|^2_{(\mathbf{I} - \mathbf{W})^\dagger} - \eta^2\|\mathbf{D}^k - \mathbf{D}^*\|^2 \\
&\quad + \frac{\gamma}{2}\mathbb{E}\|\hat{\mathbf{Z}}^{k+1} - \mathbf{Z}^{k+1}\|^2_{\mathbf{I} - \mathbf{W}} \\
&= \|\mathbf{X}^k - \mathbf{X}^* - \eta\mathbf{G}^k + \eta\nabla\mathbf{F}(\mathbf{X}^*)\|^2 + \frac{\eta^2}{\gamma}\|\mathbf{D}^k - \mathbf{D}^*\|^2_{2(\mathbf{I} - \mathbf{W})^\dagger - \gamma\mathbf{I}} + \frac{\gamma}{2}\mathbb{E}\|\hat{\mathbf{Z}}^{k+1} - \mathbf{Z}^{k+1}\|^2_{\mathbf{I} - \mathbf{W}},
\end{aligned}
$$

From the update of $\mathbf{H}$ in Line 5 of the compression procedure 'COMM', we obtain

$$
\begin{aligned}
&\mathbb{E}\|\mathbf{H}^{k+1} - \mathbf{Z}^*\|^2 \\
&= \|(1 - \alpha)(\mathbf{H}^k - \mathbf{Z}^*) + \alpha(\mathbf{Z}^{k+1} - \mathbf{Z}^*)\|^2 + \alpha^2\mathbb{E}\|\mathbf{Q}^k - \mathbb{E}\mathbf{Q}^k\|^2 \\
&= (1 - \alpha)\|\mathbf{H}^k - \mathbf{Z}^*\|^2 + \alpha\|\mathbf{Z}^{k+1} - \mathbf{Z}^*\|^2 - \alpha(1 - \alpha)\|\mathbf{Z}^{k+1} - \mathbf{H}^k\|^2 + \alpha^2\mathbb{E}\|\hat{\mathbf{Z}}^{k+1} - \mathbf{Z}^{k+1}\|^2,
\end{aligned}
$$

where the last equality uses

$$\|(1-\alpha)\mathbf{x} + \alpha\mathbf{y}\|^2 = (1-\alpha)\|\mathbf{x}\|^2 + \alpha\|\mathbf{y}\|^2 - \alpha(1-\alpha)\|\mathbf{x}-\mathbf{y}\|^2.$$

Multiplying the **H**-equality by $\sqrt{C}$ and adding it to both sides of the $(\mathbf{Z}, \mathbf{D})$-equality, we have

$$\|\mathbf{Z}^{k+1} - \mathbf{Z}^*\|^2_{(1-\sqrt{C}\alpha)\mathbf{I}-\frac{\gamma}{2}(\mathbf{I}-\mathbf{W})} + \frac{2\eta^2}{\gamma}\mathbb{E}\|\mathbf{D}^{k+1} - \mathbf{D}^*\|^2_{(\mathbf{I}-\mathbf{W})^\dagger} + \sqrt{C}\mathbb{E}\|\mathbf{H}^{k+1} - \mathbf{Z}^*\|^2$$

$$= \|\mathbf{X}^k - \mathbf{X}^* - \eta\mathbf{G}^k + \eta\nabla\mathbf{F}(\mathbf{X}^*)\|^2 + \frac{\eta^2}{\gamma}\|\mathbf{D}^k - \mathbf{D}^*\|^2_{2(\mathbf{I}-\mathbf{W})^\dagger - \gamma\mathbf{I}} + \sqrt{C}(1-\alpha)\|\mathbf{H}^k - \mathbf{Z}^*\|^2$$

$$+ \frac{1}{2}\mathbb{E}\|\hat{\mathbf{Z}}^{k+1} - \mathbf{Z}^{k+1}\|^2_{\gamma(\mathbf{I}-\mathbf{W}) + 2\sqrt{C}\alpha^2\mathbf{I}} - \sqrt{C}\alpha(1-\alpha)\mathbb{E}\|\mathbf{Z}^{k+1} - \mathbf{H}^k\|^2.$$

When $\alpha \in (0, (1+C)^{-1})$, we have $1 - \sqrt{C}\alpha \geq 1/2$. Therefore we get

$$\|\mathbf{Z}^{k+1} - \mathbf{Z}^*\|^2_{\mathbf{I}-\frac{\gamma}{2}(\mathbf{I}-\mathbf{W})} \leq M^{-1}\|\mathbf{Z}^{k+1} - \mathbf{Z}^*\|^2_{(1-\sqrt{C}\alpha)\mathbf{I}-\frac{\gamma}{2}(\mathbf{I}-\mathbf{W})},$$

when $\gamma \leq \frac{2(1-\sqrt{C}\alpha)}{\lambda_{\max}(\mathbf{I}-\mathbf{W})} \leq \frac{2}{\lambda_{\max}(\mathbf{I}-\mathbf{W})}$. Pluging it into (14), we obtain

$$M\mathbb{E}\|\mathbf{X}^{k+1} - \mathbf{X}^*\|^2 \leq \|\mathbf{Z}^{k+1} - \mathbf{Z}^*\|^2_{(1-\sqrt{C}\alpha)\mathbf{I}-\frac{\gamma}{2}(\mathbf{I}-\mathbf{W})} + M\frac{\gamma^2}{4}\mathbb{E}\|\hat{\mathbf{Z}}^{k+1} - \mathbf{Z}^{k+1}\|^2_{(\mathbf{I}-\mathbf{W})^2}.$$

Therefore

$$M\mathbb{E}\|\mathbf{X}^{k+1} - \mathbf{X}^*\|^2 + \frac{2\eta^2}{\gamma}\mathbb{E}\|\mathbf{D}^{k+1} - \mathbf{D}^*\|^2_{(\mathbf{I}-\mathbf{W})^\dagger} + \sqrt{C}\mathbb{E}\|\mathbf{H}^{k+1} - \mathbf{Z}^*\|^2$$

$$\leq \|\mathbf{X}^k - \mathbf{X}^* - \eta\mathbf{G}^k + \eta\nabla\mathbf{F}(\mathbf{X}^*)\|^2 + \frac{2\eta^2}{\gamma}\left(1 - \frac{\gamma}{2}\lambda_{\min}(\mathbf{I}-\mathbf{W})\right)\|\mathbf{D}^k - \mathbf{D}^*\|^2_{(\mathbf{I}-\mathbf{W})^\dagger}$$

$$+ \sqrt{C}(1-\alpha)\|\mathbf{H}^k - \mathbf{Z}^*\|^2 - \sqrt{C}\alpha(1-\alpha)\|\mathbf{Z}^{k+1} - \mathbf{H}^k\|^2$$

$$+ \left(M\frac{\gamma^2}{4}\lambda^2_{\max}(\mathbf{I}-\mathbf{W}) + \frac{\gamma}{2}\lambda_{\max}(\mathbf{I}-\mathbf{W}) + \sqrt{C}\alpha^2\right)\mathbb{E}\|\hat{\mathbf{Z}}^{k+1} - \mathbf{Z}^{k+1}\|^2$$

$$\leq \|\mathbf{X}^k - \mathbf{X}^* - \eta\mathbf{G}^k + \eta\nabla\mathbf{F}(\mathbf{X}^*)\|^2 + \frac{2\eta^2}{\gamma}\left(1 - \frac{\gamma}{2}\lambda_{\min}(\mathbf{I}-\mathbf{W})\right)\|\mathbf{D}^k - \mathbf{D}^*\|^2_{(\mathbf{I}-\mathbf{W})^\dagger}$$

$$+ \sqrt{C}(1-\alpha)\|\mathbf{H}^k - \mathbf{Z}^*\|^2$$

$$+ \left(M\frac{\gamma^2 C}{4}\lambda^2_{\max}(\mathbf{I}-\mathbf{W}) + \frac{\gamma C}{2}\lambda_{\max}(\mathbf{I}-\mathbf{W}) + C\sqrt{C}\alpha^2 - \sqrt{C}\alpha(1-\alpha)\right)\|\mathbf{Z}^{k+1} - \mathbf{H}^k\|^2,$$

where the second inequality follows from Assumption 2 since

$$\hat{\mathbf{Z}}^{k+1} - \mathbf{Z}^{k+1} = \mathcal{Q}(\mathbf{Z}^k - \mathbf{H}^k) - (\mathbf{Z}^k - \mathbf{H}^k).$$

Note that $M \in (0,1)$ and $\frac{\gamma}{2}\lambda_{\max}(\mathbf{I}-\mathbf{W}) < 1$, then

$$M\frac{\gamma^2}{4}\lambda^2_{\max}(\mathbf{I}-\mathbf{W})C + \frac{\gamma C}{2}\lambda_{\max}(\mathbf{I}-\mathbf{W}) + C\sqrt{C}\alpha^2 - \sqrt{C}\alpha(1-\alpha)$$

$$< \frac{\gamma^2}{4}C\lambda_{\max}(\mathbf{I}-\mathbf{W}) + \frac{\gamma C}{2}\lambda_{\max}(\mathbf{I}-\mathbf{W}) + \sqrt{C}(1+C)\alpha^2 - \sqrt{C}\alpha$$

$$< C\gamma\lambda_{\max}(\mathbf{I}-\mathbf{W}) - \sqrt{C}\Delta(\alpha)$$

$$\leq 0,$$

if $\gamma \leq \frac{\Delta(\alpha)}{\sqrt{C}\lambda_{\max}(\mathbf{I}-\mathbf{W})}$. The key inequality (15) is proved. ∎

## 4.1 The General Stochastic Setting

In the general stochastic setting with $f_i(\mathbf{x}_i) = \mathbb{E}_{\xi_i \sim \mathcal{D}_i} f_i(\mathbf{x}_i, \xi_i)$. The assumptions on $f_i$ and the variance of the stochastic gradient at the optimal point allow us to show the linear convergence of Algorithm 1 up to a neighborhood of the optimal point.

**Theorem 5 (Prox-LEAD)** *Under Assumptions 1–4, let* $\{(\mathbf{X}^k, \mathbf{D}^k, \mathbf{H}^k)\}$ *be the sequence generated by Algorithm 1 and* $\sigma^2 = \frac{1}{n}\sum_{i=1}^{n}\sigma_i^2$. *Set* $\eta \in \left(0, \frac{1}{2L}\right]$ *and* $\alpha \in \left(0, \min\left\{\frac{\eta\mu}{\sqrt{C}}, \frac{1}{1+C}\right\}\right)$ *we can choose*

$$\gamma \in \left(0, \frac{1}{\lambda_{\max}(\mathbf{I}-\mathbf{W})}\min\left\{\frac{2\eta\mu - 2\sqrt{C}\alpha}{\eta\mu}, \frac{\Delta(\alpha)}{\sqrt{C}}\right\}\right], \quad \Delta(\alpha) := \alpha - (1+C)\alpha^2 > 0.$$

*Then, in total expectation,*

$$\frac{1}{n}\mathbb{E}\Phi^{k+1} \leq \rho^k \frac{1}{n}\mathbb{E}\Phi^1 + \frac{2\eta^2\sigma^2}{1-\rho},$$

*where*

$$\Phi^k = M\|\mathbf{X}^k - \mathbf{X}^*\|^2 + \frac{2\eta^2}{\gamma}\|\mathbf{D}^k - \mathbf{D}^*\|_{(\mathbf{I}-\mathbf{W})^\dagger}^2 + \sqrt{C}\|\mathbf{H}^k - \mathbf{Z}^*\|^2$$

*and*

$$\rho = \max\left\{\frac{1-\eta\mu}{M}, 1 - \frac{\gamma}{2}\lambda_{\min}(\mathbf{I}-\mathbf{W}), 1 - \alpha\right\} < 1.$$

**Proof** In Lemma 4, we derive one-step progress inequality in expectation conditioned on the stochastic compression at $k$th step and we now focus on the term involving $\mathbf{X}$ and $\mathbf{G}^k$ in (15).

Take the conditional expectation on stochastic gradient at $k$th step, we have

$$\begin{aligned}
&\mathbb{E}\|\mathbf{X}^k - \mathbf{X}^* - \eta\mathbf{G}^k + \eta\nabla\mathbf{F}(\mathbf{X}^*)\|^2 \\
=\ &\|\mathbf{X}^k - \mathbf{X}^*\|^2 - 2\eta\langle\mathbf{X}^k - \mathbf{X}^*, \nabla\mathbf{F}(\mathbf{X}^k) - \nabla\mathbf{F}(\mathbf{X}^*)\rangle + \eta^2\mathbb{E}\|\nabla\mathbf{F}(\mathbf{X}^k, \xi^k) - \nabla\mathbf{F}(\mathbf{X}^*)\|^2 \\
\leq\ &\|\mathbf{X}^k - \mathbf{X}^*\|^2 - 2\eta\langle\mathbf{X}^k - \mathbf{X}^*, \nabla\mathbf{F}(\mathbf{X}^k) - \nabla\mathbf{F}(\mathbf{X}^*)\rangle + 2\eta^2\mathbb{E}\|\nabla\mathbf{F}(\mathbf{X}^k, \xi^k) - \nabla\mathbf{F}(\mathbf{X}^*, \xi^k)\|^2 \\
&+ 2\eta^2\mathbb{E}\|\nabla\mathbf{F}(\mathbf{X}^*, \xi^k) - \nabla\mathbf{F}(\mathbf{X}^*)\|^2 \\
\leq\ &\|\mathbf{X}^k - \mathbf{X}^*\|^2 - 2\eta\langle\mathbf{X}^k - \mathbf{X}^*, \nabla\mathbf{F}(\mathbf{X}^k) - \nabla\mathbf{F}(\mathbf{X}^*)\rangle + 4\eta^2 L \mathrm{V}_{\mathbf{F}}(\mathbf{X}^k, \mathbf{X}^*) + 2n\eta^2\sigma^2 \\
=\ &\|\mathbf{X}^k - \mathbf{X}^*\|^2 - 2\eta\langle\mathbf{X}^k - \mathbf{X}^*, \nabla\mathbf{F}(\mathbf{X}^k)\rangle + 2\eta(\mathbf{F}(\mathbf{X}^k) - \mathbf{F}(\mathbf{X}^*) - \mathrm{V}_{\mathbf{F}}(\mathbf{X}^k, \mathbf{X}^*)) \\
&+ 4\eta^2 L\mathrm{V}_{\mathbf{F}}(\mathbf{X}^k, \mathbf{X}^*) + 2n\eta^2\sigma^2 \\
=\ &\|\mathbf{X}^k - \mathbf{X}^*\|^2 - 2\eta(\mathbf{F}(\mathbf{X}^*) - \mathbf{F}(\mathbf{X}^k) - \langle\mathbf{X}^* - \mathbf{X}^k, \nabla\mathbf{F}(\mathbf{X}^k)\rangle) + 2n\eta^2\sigma^2 \\
&- 2\eta(1 - 2\eta L)\mathrm{V}_{\mathbf{F}}(\mathbf{X}^k, \mathbf{X}^*) \\
\leq\ &(1 - \eta\mu)\|\mathbf{X}^k - \mathbf{X}^*\|^2 + 2n\eta^2\sigma^2,
\end{aligned}$$

where the first equality uses the unbiasedness of stochastic gradient, the second inequality follows the expected Lipschitz property in Assumption 4, and the last inequality is due to the strong convexity and $\eta \leq \frac{1}{2L}$.

Now we use the power property to take the conditional expectation on the stochastic gradient for (15) and plug the above inequality into it, then

$$
\begin{aligned}
M\mathbb{E}\|\mathbf{X}^{k+1} &- \mathbf{X}^*\|^2 + \frac{2\eta^2}{\gamma}\mathbb{E}\|\mathbf{D}^{k+1} - \mathbf{D}^*\|_{(\mathbf{I}-\mathbf{W})^\dagger}^2 + \sqrt{C}\mathbb{E}\|\mathbf{H}^{k+1} - \mathbf{Z}^*\|^2 \\
\leq &\ (1 - \eta\mu)\|\mathbf{X}^k - \mathbf{X}^*\|^2 + \frac{2\eta^2}{\gamma}\left(1 - \frac{\gamma}{2}\lambda_{\min}(\mathbf{I}-\mathbf{W})\right)\|\mathbf{D}^k - \mathbf{D}^*\|_{(\mathbf{I}-\mathbf{W})^\dagger}^2 \\
&+ \sqrt{C}(1-\alpha)\|\mathbf{H}^k - \mathbf{Z}^*\|^2 + 2n\eta^2\sigma^2.
\end{aligned}
\tag{16}
$$

Define

$$
\Phi^k = M\|\mathbf{X}^k - \mathbf{X}^*\|^2 + \frac{2\eta^2}{\gamma}\|\mathbf{D}^k - \mathbf{D}^*\|_{(\mathbf{I}-\mathbf{W})^\dagger}^2 + \sqrt{C}\|\mathbf{H}^k - \mathbf{Z}^*\|^2,
$$

by (16), we have

$$
\mathbb{E}\Phi^{k+1} \leq \max\left\{\frac{1-\eta\mu}{M}, 1 - \frac{\gamma}{2}\lambda_{\min}(\mathbf{I}-\mathbf{W}), 1-\alpha\right\}\Phi^k + 2n\eta^2\sigma^2.
$$

Note that $\gamma \leq \frac{2\eta\mu - 2\sqrt{C}\alpha}{\lambda_{\max}(\mathbf{I}-\mathbf{W})\eta\mu} < \frac{2(1-\sqrt{C}\alpha)}{\lambda_{\max}(\mathbf{I}-\mathbf{W})}$ and $\alpha < \frac{\mu\eta}{\sqrt{C}}$ guarantee

$$
\frac{1-\eta\mu}{M} = \frac{1-\eta\mu}{1 - \frac{\sqrt{C}\alpha}{1 - \frac{\gamma}{2}\lambda_{\max}(\mathbf{I}-\mathbf{W})}} < 1.
$$

Finally, by taking the total expectation on both sides of the above inequality, the proof is complete with the given choice of parameters. ∎

When there is no compression, Prox-LEAD is reduced to the special case of PUDA with stochastic gradient. Corollary 6 shows the linear convergence to the neighborhood of the optimal solution and when the gradient is deterministic, the convergence rate matches that given in Alghunaim et al. (2020).

**Corollary 6 (Stochastic PUDA)** *When there is no compression, i.e., $C = 0$, under Assumptions 1, 3 and 4, we can pick $\alpha = 1$ and $\gamma = 1$. Then, for any $\eta \in \left(0, \frac{1}{2L}\right]$*

$$
\begin{aligned}
\mathbb{E}\|\mathbf{X}^{k+1} &- \mathbf{X}^*\|^2 + 2\eta^2\mathbb{E}\|\mathbf{D}^{k+1} - \mathbf{D}^*\|^2 \\
\leq &\ \max\left\{1 - \eta\mu, 1 - \frac{\lambda_{\min}(\mathbf{I}-\mathbf{W})}{2}\right\}\left(\mathbb{E}\|\mathbf{X}^k - \mathbf{X}^*\|^2 + 2\eta^2\mathbb{E}\|\mathbf{D}^k - \mathbf{D}^*\|^2\right) + 2n\eta^2\sigma^2.
\end{aligned}
\tag{17}
$$

**Proof** When $C = 0$, we get $M = 1$ by its definition. We can take $\alpha = 1 - \epsilon$ for some $\epsilon \in (0, 1)$. Note that $\frac{\Delta(\alpha)}{\sqrt{C}}$ approaches infinity as $C$ goes to 0, the upper bound of $\gamma$ is reduced to $\frac{2}{\lambda_{\max}(\mathbf{I}-\mathbf{W})}$, which is strictly greater than 1 due to Assumption 1. Hence we can take $\gamma = 1$ and the convergence rate becomes

$$\max\left\{1 - \eta\mu, 1 - \frac{\lambda_{\min}(\mathbf{I}-\mathbf{W})}{2}, \epsilon\right\}.$$

Let $\epsilon$ approach 0 then we complete the proof. ∎

The next theorem shows Prox-LEAD has $\mathcal{O}(1/k)$ convergence rate to the exact optimal solution. Before presenting the convergence results, we define

$$\kappa_f = \frac{L}{\mu}, \quad \kappa_g = \frac{\lambda_{\max}(\mathbf{I}-\mathbf{W})}{\lambda_{\min}(\mathbf{I}-\mathbf{W})}$$

as the condition numbers of the objective function and the network, respectively.

**Theorem 7 (Diminishing stepsize)** *Under Assumptions 1–4, let $\alpha^k = \frac{\eta^k\mu}{1+C}$ and $\gamma^k = \frac{\eta^k\mu}{2(1+C)^2\lambda_{\max}(\mathbf{I}-\mathbf{W})}$ with $\eta^k = \frac{8(1+C)^2\kappa_g\kappa_f}{k+16(1+C)^2\kappa_g\kappa_f}\frac{1}{L}$, we have*

$$\frac{1}{n}\sum_{i=1}^{n}\mathbb{E}\|\mathbf{x}_i^k - \mathbf{x}^*\|^2 \leq \frac{\max\{16(1+C)^2\kappa_f\kappa_g\frac{\Phi^0}{n}, \frac{128\sigma^2(1+C)^4\kappa_f^2\kappa_g^2}{L^2}\}}{k+16(1+C)^2\kappa_g\kappa_f}\frac{1}{\widetilde{M}},$$

*where $\widetilde{M} := 1 - \frac{2\sqrt{C}}{3(1+C)\kappa_f} \in [\frac{2}{3}, 1]$.*

*In particular, when there is no compression, i.e., $C = 0$, we have $\widetilde{M} = 1$ and*

$$\frac{1}{n}\sum_{i=1}^{n}\mathbb{E}\|\mathbf{x}_i^k - \mathbf{x}^*\|^2 \leq \frac{\max\{16\kappa_f\kappa_g\frac{\Phi^0}{n}, \frac{128\sigma^2\kappa_f^2\kappa_g^2}{L^2}\}}{k+16\kappa_g\kappa_f}.$$

**Proof** We use the descent inequality in Theorem 5 to show the sublinear convergence. Firstly, we verify that the given parameters are feasible.

**Parameter feasibility** Notice that $\eta^k\mu \in \left(0, \frac{\mu}{2L}\right] \subset \left(0, \frac{1}{2}\right]$ for $k \geq 0$ and $\frac{\sqrt{C}}{1+C} \in \left[0, \frac{1}{2}\right]$, then

$$\alpha^k = \frac{\eta^k\mu}{1+C} < \min\left\{\frac{\eta^k\mu}{\sqrt{C}}, \frac{1}{1+C}\right\},$$

which shows $\alpha^k$ is in the feasible region.

Plugging $\alpha^k$ into the upper bound of $\gamma$, we have

$$\frac{1}{\lambda_{\max}(\mathbf{I}-\mathbf{W})}\min\left\{2 - \frac{2\sqrt{C}}{1+C}, \frac{\eta^k\mu - (\eta^k)^2\mu^2}{(1+C)\sqrt{C}}\right\} \geq \frac{\eta^k\mu}{(1+C)^2\lambda_{\max}(\mathbf{I}-\mathbf{W})}$$

since $2 - \frac{2\sqrt{C}}{1+C} \geq 1$ and $\frac{\eta^k\mu-(\eta^k)^2\mu^2}{(1+C)\sqrt{C}} \geq \frac{2(\eta^k\mu-(\eta^k)^2\mu^2)}{(1+C)^2} \geq \frac{\eta^k\mu}{(1+C)^2}$.

Hence $\gamma^k = \frac{\eta^k\mu}{2(1+C)^2\lambda_{\max}(\mathbf{I}-\mathbf{W})}$ is feasible.

**Sublinear convergence**   Now the inequality in Theorem 5 gives

$$\mathbb{E}\Phi^{k+1} \leq \max\{\rho_1, \rho_2, \rho_3\}\mathbb{E}\Phi^k + 2n(\eta^k)^2\sigma^2$$

with

$$\rho_1 := \frac{1-\eta^k\mu}{1-\frac{\sqrt{C}\alpha}{1-\frac{\gamma^k}{2}\lambda_{\max}(\mathbf{I}-\mathbf{W})}} = \frac{(1-\eta^k\mu)\left(1-\frac{\eta^k\mu}{2(1+C)^2}\right)}{1-\frac{\eta^k\mu}{2(1+C)^2}-\frac{\sqrt{C}\eta^k\mu}{(1+C)}} \leq 1-\mu\left(1-\frac{\sqrt{C}}{1+C}\right)\eta =: 1-a_1\eta^k,$$

$$\rho_2 := 1-\frac{\gamma^k}{2}\lambda_{\min}(\mathbf{I}-\mathbf{W}) = 1-\frac{\mu}{4(1+C)^2\kappa_g}\eta^k =: 1-a_2\eta^k,$$

$$\rho_3 := 1-\alpha^k = 1-\frac{\mu}{1+C}\eta^k =: 1-a_3\eta^k.$$

To verify the inequality of $\rho_1$, it's equivalent to verify

$$(1-\eta^k\mu)\left(1-\frac{\eta^k\mu}{2(1+C)^2}\right) \leq \left[1-\left(1-\frac{\sqrt{C}}{1+C}\right)\mu\eta^k\right]\left[1-\frac{\eta^k\mu}{2(1+C)^2}-\frac{\sqrt{C}\eta^k\mu}{(1+C)}\right].$$

Let $a = 1-\eta^k\mu, b = 1-\frac{\eta^k\mu}{2(1+C)^2}$ and $x = \frac{\sqrt{C}}{1+C}\eta^k\mu$, then the inequality is equivalent to $ab \leq (a+x)(b-x) = ab + (b-a)x - x^2$. Notice that this holds when $0 \leq x \leq b-a$, then we only need to verify $x \leq b-a$, which is clear due to $\frac{\sqrt{C}}{1+C} \in [0, \frac{1}{2}]$ and

$$x = \frac{\sqrt{C}}{1+C}\eta^k\mu \leq \frac{\eta^k\mu}{2} \leq \eta^k\mu\left(1-\frac{1}{2(1+C)^2}\right) = b-a.$$

Note that $a_2 = \min\{a_1, a_2, a_3\}$, we get the simplified one-step progress inequality,

$$\mathbb{E}\Phi^{k+1} \leq (1-a_2\eta^k)\mathbb{E}\Phi^k + 2n(\eta^k)^2\sigma^2, \tag{18}$$

and we will use it to show the sublinear convergence by induction.

Let $A := \frac{8(1+C)^2\kappa_g}{\mu}, B := 16(1+C)^2\kappa_f\kappa_g$ and $D := \max\{16(1+C)^2\kappa_f\kappa_g\Phi^0, \frac{128n\sigma^2(1+C)^4\kappa_f^2\kappa_g^2}{L^2}\}$, we claim that

$$\mathbb{E}\Phi^k \leq \frac{D}{k+B} \tag{19}$$

with $\eta^k = \frac{A}{k+B}$.

- Firstly, when $k = 0$,

$$\frac{D}{B} = \max\left\{\Phi^0, \frac{8n\sigma^2(1+C)^2\kappa_f\kappa_g}{L^2}\right\} \geq \Phi^0.$$

- Suppose the claim holds for $k > 0$, from the inequality (18), we have

$$\mathbb{E}\Phi^{k+1} \leq \left(1-a_2\frac{A}{k+B}\right)\frac{D}{k+B} + 2n\sigma^2\left(\frac{A}{k+B}\right)^2.$$

Multiplying $(k + B)(k + 1 + B)$ on both sides, we get

$$
\begin{aligned}
&(k + B)(k + 1 + B)\mathbb{E}\Phi^{k+1} \\
&\leq \left(1 - a_2\frac{A}{k+B}\right)D(k+1+B) + \frac{2n\sigma^2 A^2(k+1+B)}{k+B} \\
&= \frac{D(k+B)^2 + D(k+B) - a_2 AD(k+1+B) + 2n\sigma^2 A^2(k+1+B)}{k+B} \\
&= D(k+B) + (D + 2n\sigma^2 A^2 - a_2 AD) + \frac{2n\sigma^2 A^2 - a_2 AD}{k+B}.
\end{aligned}
$$

By the definition of $A$ and $\mathrm{D}$, we have $a_2 A = 2$ and

$$
D \geq \frac{128n\sigma^2(1+C)^4\kappa_f^2\kappa_g^2}{L^2} = 2n\sigma^2 A^2,
$$

then $(k + B)(k + 1 + B)\mathbb{E}\Phi^{k+1} \leq D(k + B)$, which shows

$$
\mathbb{E}\Phi^{k+1} \leq \frac{D}{k+1+B}.
$$

Finally, from the definition of $\Phi^k$, we have

$$
M\sum_{i=1}^{n}\mathbb{E}\|\mathbf{x}_i^k - \mathbf{x}^*\|^2 \leq \mathbb{E}\Phi^k,
$$

where $M$ is a decreasing function of $\eta$. Then

$$
\begin{aligned}
M = 1 - \frac{\frac{\sqrt{C}}{1+C}\eta\mu}{1 - \frac{\eta\mu}{2(1+C)^2}} &\geq 1 - \frac{\frac{\sqrt{C}}{1+C}\frac{1}{2\kappa_f}}{1 - \frac{1}{4(1+C)^2\kappa_f}} \\
&= 1 - \frac{2\sqrt{C}(1+C)}{4(1+C)^2\kappa_f - 1} \geq 1 - \frac{2\sqrt{C}}{3(1+C)\kappa_f} = \widetilde{M}.
\end{aligned}
$$

Multiplying $\widetilde{M}^{-1}$ on both sides of (19) and the proof is complete. ∎

Theorem 7 uses diminishing stepsizes to establish the exact convergence to the optimal solution while the convergence rate is only guaranteed to be sublinear. In order to maintain the linear convergence, we consider the problems with finite-sum structure and adapt Prox-LEAD with two variance reduction schemes: Loopless SVRG and SAGA.

## 4.2 Finite-sum Setting with Variance Reduction

In the finite-sum setting, we assume that each $f_i$ is the average of $m$ functions $f_{ij}$ and impose the commonly assumed $L$-Lipschitz continuity to $\nabla f_{ij}$. We will keep using $\Phi^k$ and $M$ defined in Theorem 5.

**Theorem 8 (Loopless SVRG)** *Under Assumptions 1, 2, and 4, for any $p \in (0,1)$, set $p_{ij} = \frac{1}{m}, \forall i \in [n], \forall j \in [m]$, $\eta = \frac{1}{6L}, \alpha = \frac{1}{12(1+C)\kappa_f}$, and*

$$\gamma = \min\left\{ \frac{1}{24\sqrt{C}(1+C)\lambda_{\max}(\mathbf{I}-\mathbf{W})\kappa_f}, \frac{1}{24(1+C)\lambda_{\max}(\mathbf{I}-\mathbf{W})} \right\}.$$

*Then we have*

$$\mathbb{E}\widetilde{\Phi}^{k+1} \leq \left(1 - \left(\max\left\{48\sqrt{C}(1+C)\kappa_f\kappa_g, 12(1+C)\kappa_f, \frac{282\kappa_f}{23}, 48(1+C)\kappa_g, \frac{2}{p}\right\}\right)^{-1}\right)^{k+1}\widetilde{\Phi}^0,$$

*where*

$$\widetilde{\Phi}^k := \Phi^k + \frac{2}{9pL}\sum_{i=1}^n \mathrm{V}_{f_i}(\widetilde{\mathbf{x}}_i^k, \mathbf{x}^*).$$

## Proof

**Linear convergence**  The following proof is applicable to the general choice of $\{p_{ij}\}$ and for simplicity, we only consider the uniform sampling case with $p_{ij} = \frac{1}{m}$.

Lemma 4 still holds with different $\mathbf{G}^k = [\mathbf{g}_1^k, \cdots, \mathbf{g}_n^k]^\top$ in procedure SGO, then we focus on the following term

$$\mathbb{E}\|\mathbf{X}^k - \mathbf{X}^* - \eta\mathbf{G}^k + \eta\nabla\mathbf{F}(\mathbf{X}^*)\|^2$$

$$= \sum_{i=1}^n \mathbb{E}\left\|\mathbf{x}_i^k - \mathbf{x}^* - \eta\left(\frac{\nabla f_{il}(\mathbf{x}_i^k) - \nabla f_{il}(\widetilde{\mathbf{x}}_i^k)}{mp_{il}} + \nabla f_i(\widetilde{\mathbf{x}}_i^k) - \nabla f_i(\mathbf{x}^*)\right)\right\|^2$$

$$= \|\mathbf{X}^k - \mathbf{X}^*\|^2 - 2\eta\sum_{i=1}^n \mathbb{E}\left\langle \mathbf{x}_i^k - \mathbf{x}^*, \frac{\nabla f_{il}(\mathbf{x}_i^k) - \nabla f_{il}(\widetilde{\mathbf{x}}_i^k)}{mp_{il}} + \nabla f_i(\widetilde{\mathbf{x}}_i^k) - \nabla f_i(\mathbf{x}^*)\right\rangle$$

$$\quad + \eta^2\sum_{i=1}^n \mathbb{E}\left\|\frac{\nabla f_{il}(\mathbf{x}_i^k) - \nabla f_{il}(\widetilde{\mathbf{x}}_i^k)}{mp_{il}} + \nabla f_i(\widetilde{\mathbf{x}}_i^k) - \nabla f_i(\mathbf{x}^*)\right\|^2$$

$$= \|\mathbf{X}^k - \mathbf{X}^*\|^2 - 2\eta\sum_{i=1}^n \langle\mathbf{x}_i^k - \mathbf{x}^*, \nabla f_i(\mathbf{x}_i^k) - \nabla f_i(\mathbf{x}^*)\rangle$$

$$\quad + \eta^2\sum_{i=1}^n\sum_{j=1}^m p_{ij}\left\|\frac{\nabla f_{ij}(\mathbf{x}_i^k) - \nabla f_{ij}(\widetilde{\mathbf{x}}_i^k)}{mp_{ij}} + \nabla f_i(\widetilde{\mathbf{x}}_i^k) - \nabla f_i(\mathbf{x}^*)\right\|^2$$

$$= \|\mathbf{X}^k - \mathbf{X}^*\|^2 - 2\eta\sum_{i=1}^n \langle\mathbf{x}_i^k - \mathbf{x}^*, \nabla f_i(\mathbf{x}_i^k)\rangle + 2\eta\sum_{i=1}^n (f_i(\mathbf{x}_i^k) - f_i(\mathbf{x}^*) - \mathrm{V}_{f_i}(\mathbf{x}_i^k, \mathbf{x}^*))$$

$$\quad + \eta^2\sum_{i=1}^n\sum_{j=1}^m p_{ij}\left\|\frac{\nabla f_{ij}(\mathbf{x}_i^k) - \nabla f_{ij}(\mathbf{x}^*) + f_{ij}(\mathbf{x}^*) - \nabla f_{ij}(\widetilde{\mathbf{x}}_i^k)}{mp_{ij}} + \nabla f_i(\widetilde{\mathbf{x}}_i^k) - \nabla f_i(\mathbf{x}^*)\right\|^2$$

$$\leq (1-\mu\eta)\|\mathbf{X}^k - \mathbf{X}^*\| - 2\eta\sum_{i=1}^n \mathrm{V}_{f_i}(\mathbf{x}_i^k, \mathbf{x}^*) + 2\eta^2\sum_{i=1}^n\sum_{j=1}^m \frac{1}{m^2p_{ij}}\|\nabla f_{ij}(\mathbf{x}_i^k) - \nabla f_{ij}(\mathbf{x}^*)\|^2$$

$$\quad + 2\eta^2\sum_{i=1}^n\sum_{j=1}^m p_{ij}\left\|\frac{\nabla f_{ij}(\mathbf{x}^*) - \nabla f_{ij}(\widetilde{\mathbf{x}}_i^k)}{mp_{ij}} + \nabla f_i(\widetilde{\mathbf{x}}_i^k) - \nabla f_i(\mathbf{x}^*)\right\|^2,$$

where the inequality uses the strong convexity of $f_{ij}$ in Assumption 4 and $(a+b)^2 \leq 2a^2 + 2b^2$.

Let $u_i$ be the random variable taking values in $\left\{ \frac{1}{mp_{il}} (\nabla f_{il}(\tilde{\mathbf{x}}_i^k) - \nabla f_{il}(\mathbf{x}^*)) : l \in [m] \right\}$ with distribution $\mathcal{P}_i = \{p_{il} : l \in [m]\}$, then the last term is actually the summation of the variance of $u_i$s due to

$$\mathbb{E}u_i = \sum_{j=1}^m p_{ij} \frac{1}{mp_{ij}} (\nabla f_{ij}(\tilde{\mathbf{x}}_i^*) - \nabla f_{ij}(\mathbf{x}^*)) = \nabla f_i(\tilde{\mathbf{x}}_i^k) - \nabla f_i(\mathbf{x}^*).$$

Applying the inequality $\mathbb{E}\|u_i - \mathbb{E}u_i\|^2 \leq \mathbb{E}\|u_i\|^2$ to the last term, we get

$$\begin{aligned}
&\mathbb{E}\|\mathbf{X}^k - \mathbf{X}^* - \eta\mathbf{G}^k + \eta\nabla\mathbf{F}(\mathbf{X}^*)\|^2 \\
&\leq (1-\mu\eta)\|\mathbf{X}^k - \mathbf{X}^*\| - 2\eta\sum_{i=1}^n \mathrm{V}_{f_i}(\mathbf{x}_i^k, \mathbf{x}^*) + \frac{4\eta^2 L}{mp_{\min}}\sum_{i=1}^n \mathrm{V}_{f_i}(\mathbf{x}_i^k, \mathbf{x}^*) \\
&\quad + 2\eta^2 \sum_{i=1}^n \sum_{j=1}^m \frac{1}{m^2 p_{ij}} \|\nabla f_{ij}(\tilde{\mathbf{x}}_i^k) - \nabla f_{ij}(\mathbf{x}^*)\|^2 \\
&\leq (1-\mu\eta)\|\mathbf{X}^k - \mathbf{X}^*\| - 2\eta\sum_{i=1}^n \mathrm{V}_{f_i}(\mathbf{x}_i^k, \mathbf{x}^*) + \frac{4\eta^2 L}{mp_{\min}}\sum_{i=1}^n \mathrm{V}_{f_i}(\mathbf{x}_i^k, \mathbf{x}^*) \\
&\quad + \frac{4\eta^2 L}{mp_{\min}}\sum_{i=1}^n \mathrm{V}_{f_i}(\tilde{\mathbf{x}}_i^k, \mathbf{x}^*),
\end{aligned}$$

where $p_{\min} := \min_{i,j}\{p_{ij}\}$ and the first inequality uses the Lipschitz smoothness of $f_{ij}$ in Assumption 4.

From the update of $\tilde{\mathbf{x}}_i^{k+1}$, we have

$$\mathbb{E}\tilde{\mathbf{x}}_i^{k+1} = p\mathbf{x}_i^k + (1-p)\tilde{\mathbf{x}}_i^k.$$

Hence

$$\mathbb{E}\mathrm{V}_{f_i}(\tilde{\mathbf{x}}_i^{k+1}, \mathbf{x}^*) = p\mathrm{V}_{f_i}(\mathbf{x}_i^k, \mathbf{x}^*) + (1-p)\mathrm{V}_{f_i}(\tilde{\mathbf{x}}_i^k, \mathbf{x}^*).$$

Combine them with the inequality from Lemma 4, we get

$$\begin{aligned}
&\mathbb{E}\Phi^{k+1} + \tilde{c}\sum_{i=1}^n \mathbb{E}\mathrm{V}_{f_i}(\tilde{\mathbf{x}}_i^{k+1}, \mathbf{x}^*) \\
&\leq (1-\eta\mu)\|\mathbf{X}^k - \mathbf{X}^*\|^2 - \left(2\eta - \frac{4\eta^2 L}{mp_{\min}} - \tilde{c}p\right)\sum_{i=1}^n \mathrm{V}_{f_i}(\mathbf{x}_i^k, \mathbf{x}^*) \\
&\quad + \frac{4\eta^2 L}{mp_{\min}}\sum_{i=1}^n \mathrm{V}_{f_i}(\tilde{\mathbf{x}}_i^k, \mathbf{x}^*) + \frac{2\eta^2}{\gamma}\left(1 - \frac{\gamma}{2}\lambda_{\min}(\mathbf{I} - \mathbf{W})\right)\|\mathbf{D}^k - \mathbf{D}^*\|_{(\mathbf{I}-\mathbf{W})^\dagger}^2 \\
&\quad + \sqrt{C}(1-\alpha)\|\mathbf{H}^k - \mathbf{Z}^*\|^2 + \tilde{c}(1-p)\sum_{i=1}^n \mathrm{V}_{f_i}(\tilde{\mathbf{x}}_i^k, \mathbf{x}^*),
\end{aligned}$$

where $\tilde{c} = \frac{8\eta^2 L}{pmp_{\min}}$.

By the choice of $\eta$ and $\{p_{ij}\}$, we have

$$\frac{4\eta^2 L}{mp_{\min}} + \tilde{c}p = \frac{1}{18L}(2+4) = \frac{1}{3L} = 2\eta.$$

Therefore,

$$\mathbb{E}\Phi^{k+1} + \tilde{c}\sum_{i=1}^{n}\mathbb{E}\mathrm{V}_{f_i}(\tilde{\mathbf{x}}_i^{k+1},\mathbf{x}^*)$$

$$\leq (1-\eta\mu)\|\mathbf{X}^k - \mathbf{X}^*\|^2 + \frac{2\eta^2}{\gamma}\left(1 - \frac{\gamma}{2}\lambda_{\min}(\mathbf{I}-\mathbf{W})\right)\|\mathbf{D}^k - \mathbf{D}^*\|^2_{(\mathbf{I}-\mathbf{W})^\dagger}$$

$$+ \sqrt{C}(1-\alpha)\|\mathbf{H}^k - \mathbf{Z}^*\|^2 + \tilde{c}\left(1 - p + \frac{4\eta^2 L}{\tilde{c}mp_{\min}}\right)\sum_{i=1}^{n}\mathrm{V}_{f_i}(\tilde{\mathbf{x}}_i^k,\mathbf{x}^*).$$

Note that

$$1 - p + \frac{4\eta^2 L}{\tilde{c}mp_{\min}} = \frac{p}{2},$$

then we get

$$\mathbb{E}\Phi^{k+1} + \frac{2}{9pL}\sum_{i=1}^{n}\mathbb{E}\mathrm{V}_{f_i}(\tilde{\mathbf{x}}_i^{k+1},\mathbf{x}^*)$$

$$\leq \max\left\{\frac{1-\eta\mu}{M}, 1 - \frac{\gamma}{2}\lambda_{\min}(\mathbf{I}-\mathbf{W}), 1-\alpha, 1 - \frac{p}{2}\right\}\left(\Phi^k + \frac{2}{9pL}\sum_{i=1}^{n}\mathrm{V}_{f_i}(\tilde{\mathbf{x}}_i^k,\mathbf{x}^*)\right).$$

**Complexity**  The linear convergence requires

$$\alpha < \min\left\{\frac{\eta\mu}{\sqrt{C}}, \frac{1}{1+C}\right\},$$

$$\gamma \leq \min\left\{\frac{2\eta\mu - 2\sqrt{C}\alpha}{\lambda_{\max}(\mathbf{I}-\mathbf{W})\eta\mu}, \frac{(1+C)\alpha^2 - \alpha}{\lambda_{\max}(\mathbf{I}-\mathbf{W})\sqrt{C}}\right\}.$$

Take $\alpha = \frac{1}{12(1+C)\kappa_f} < \frac{1}{2(1+C)}$ which is clearly in the feasible region of $\alpha$ and

$$\gamma = \min\left\{\frac{1}{24\sqrt{C}(1+C)\lambda_{\max}(\mathbf{I}-\mathbf{W})\kappa_f}, \frac{1}{24(1+C)\lambda_{\max}(\mathbf{I}-\mathbf{W})}\right\},$$

then there are two cases to be considered:

- $C \geq \frac{1}{\kappa_f^2}$ :

  In this case, $\gamma = \frac{1}{24\sqrt{C}(1+C)\lambda_{\max}(\mathbf{I}-\mathbf{W})\kappa_f}$ and it can be verified that

$$\frac{1}{24\sqrt{C}(1+C)\lambda_{\max}(\mathbf{I}-\mathbf{W})\kappa_f} \leq \min\left\{\frac{2 - \frac{\sqrt{C}}{6(1+C)}}{\lambda_{\max}(\mathbf{I}-\mathbf{W})}, \frac{1}{24\lambda_{\max}(\mathbf{I}-\mathbf{W})\sqrt{C}(1+C)\kappa_f}\right\}.$$

Since $\sqrt{C} \leq 1 + C$ when $C \geq 0$, we have the following estimation on the convergence rate

$$M = 1 - \frac{\sqrt{C}\alpha}{1 - \frac{\gamma}{2}\lambda_{\max}(\mathbf{I} - \mathbf{W})}$$

$$= 1 - \frac{\frac{\sqrt{C}}{12(1+C)\kappa_f}}{1 - \frac{1}{48\sqrt{C}(1+C)\kappa_f}} \geq 1 - \frac{4\sqrt{C}}{47(1+C)\kappa_f},$$

$$\frac{1 - \eta\mu}{M} \leq 1 - \frac{\frac{1}{6\kappa_f} - \frac{4\sqrt{C}}{47(1+C)\kappa_f}}{1 - \frac{4\sqrt{C}}{47(1+C)\kappa_f}}$$

$$\leq 1 - \frac{47(1+C) - 24\sqrt{C}}{282(1+C)\kappa_f - 24\sqrt{C}}$$

$$\leq 1 - \frac{23(1+C)}{282(1+C)\kappa_f - 24(1+C)} < 1 - \frac{23}{282\kappa_f},$$

$$1 - \frac{\gamma}{2}\lambda_{\min}(\mathbf{I} - \mathbf{W}) = 1 - \frac{1}{48\sqrt{C}(1+C)\kappa_g\kappa_f}.$$

- $C < \frac{1}{\kappa_f^2}$:

  In this case, $\gamma = \frac{1}{24(1+C)\lambda_{\max}(\mathbf{I}-\mathbf{W})}$ and it can be verified that

  $$\frac{1}{24(1+C)\lambda_{\max}(\mathbf{I}-\mathbf{W})} \leq \min\left\{\frac{2 - \frac{\sqrt{C}}{6(1+C)}}{\lambda_{\max}(\mathbf{I}-\mathbf{W})}, \frac{1}{24\lambda_{\max}(\mathbf{I}-\mathbf{W})\sqrt{C}(1+C)\kappa_f}\right\}.$$

  We have

  $$M = 1 - \frac{\sqrt{C}\alpha}{1 - \frac{\gamma}{2}\lambda_{\max}(\mathbf{I} - \mathbf{W})}$$

  $$= 1 - \frac{\frac{\sqrt{C}}{12(1+C)\kappa_f}}{1 - \frac{1}{48(1+C)}} \geq 1 - \frac{4\sqrt{C}}{47(1+C)\kappa_f},$$

  $$\frac{1 - \eta\mu}{M} = 1 - \frac{\frac{1}{6\kappa_f} - \frac{4\sqrt{C}}{47(1+C)\kappa_f}}{1 - \frac{4\sqrt{C}}{47(1+C)\kappa_f}} < 1 - \frac{23}{282\kappa_f},$$

  $$1 - \frac{\gamma}{2}\lambda_{\max}(\mathbf{I} - \mathbf{W}) = 1 - \frac{1}{48(1+C)\kappa_g}.$$

  Therefore, we have

  $$\mathbb{E}\widetilde{\Phi}^{k+1} \leq \left(1 - \left(\max\left\{48\sqrt{C}(1+C)\kappa_f\kappa_g, 12(1+C)\kappa_f, \frac{282\kappa_f}{23}, 48(1+C)\kappa_g, \frac{2}{p}\right\}\right)^{-1}\right)\widetilde{\Phi}^k$$

  and the proof is complete. ∎

**Theorem 9 (SAGA)** *Under Assumption 1, 2 and 4, set $p_{ij} = \frac{1}{m}, \forall i \in [n], \forall j \in [m]$, $\eta = \frac{1}{6L}, \alpha = \frac{1}{12(1+C)\kappa_f}$ and*

$$\gamma = \min\left\{\frac{1}{24\sqrt{C}(1+C)\lambda_{\max}(\mathbf{I}-\mathbf{W})\kappa_f}, \frac{1}{24(1+C)\lambda_{\max}(\mathbf{I}-\mathbf{W})}\right\},$$

*we will reach $\frac{1}{n}\mathbb{E}\|\mathbf{X}^k - \mathbf{X}^*\|^2 \leq \epsilon$ in total expectation after the number of iterations*

$$\mathbb{E}\widetilde{\Phi}^{k+1} \leq \left(1 - \min\left\{\frac{1}{48\sqrt{C}(1+C)\kappa_g\kappa_f}, \frac{1}{12(1+C)\kappa_f}, \frac{23}{282\kappa_f}, \frac{1}{48(1+C)\kappa_g}, \frac{1}{2m}\right\}\right)^{k+1}\widetilde{\Phi}^0,$$

*where*

$$\widetilde{\Phi}^k = \Phi^k + \frac{2}{9L}\sum_{i=1}^{n}\sum_{j=1}^{m}\mathrm{V}_{f_{ij}}(\widetilde{\mathbf{x}}_{ij}^k, \mathbf{x}^*).$$

## Proof

**Linear convergence** The following proof can be adapted to to show the convergence with the general $\{p_{ij}\}$ while for simplicity, we focus on uniform sampling case.

We start from the gradient term of the key inequality in Lemma 4 and replace $\nabla \mathbf{F}(\mathbf{X}^k, \xi^k)$ by $\mathbf{G}^k = [\mathbf{g}_1^k, \cdots, \mathbf{g}_n^k]^\top$,

$$\mathbb{E}\|\mathbf{X}^k - \mathbf{X}^* - \eta\mathbf{G}^k + \eta\nabla\mathbf{F}(\mathbf{X}^*)\|^2$$

$$= \sum_{i=1}^{n}\mathbb{E}\left\|\mathbf{x}_i^k - \mathbf{x}^* - \eta\left(\frac{\nabla f_{il}(\mathbf{x}_i^k) - \nabla f_{il}(\widetilde{\mathbf{x}}_{il}^k)}{mp_{il}} + \frac{\sum_{j=1}^{m}\nabla f_{ij}(\widetilde{\mathbf{x}}_{ij}^k) - \nabla f_{ij}(\mathbf{x}^*)}{m}\right)\right\|^2$$

$$= \|\mathbf{X}^k - \mathbf{X}^*\|^2 - 2\eta\sum_{i=1}^{n}\mathbb{E}\left\langle \mathbf{x}_i^k - \mathbf{x}^*, \frac{\nabla f_{il}(\mathbf{x}_i^k) - \nabla f_{il}(\widetilde{\mathbf{x}}_{il}^k)}{mp_{il}} + \frac{\sum_{j=1}^{m}\nabla f_{ij}(\widetilde{\mathbf{x}}_{ij}^k) - \nabla f_{ij}(\mathbf{x}^*)}{m}\right\rangle$$

$$+ \eta^2\sum_{i=1}^{n}\mathbb{E}\left\|\frac{\nabla f_{il}(\mathbf{x}_i^k) - \nabla f_{il}(\widetilde{\mathbf{x}}_{il}^k)}{mp_{il}} + \frac{\sum_{j=1}^{m}\nabla f_{ij}(\widetilde{\mathbf{x}}_{ij}^k) - \nabla f_{ij}(\mathbf{x}^*)}{m}\right\|^2$$

$$= \|\mathbf{X}^k - \mathbf{X}^*\|^2 - 2\eta\sum_{i=1}^{n}\langle\mathbf{x}_i^k - \mathbf{x}^*, \nabla f_i(\mathbf{x}_i^k) - \nabla f_i(\mathbf{x}^*)\rangle$$

$$+ \eta^2\sum_{i=1}^{n}\sum_{j=1}^{m}p_{ij}\left\|\frac{\nabla f_{ij}(\mathbf{x}_i^k) - \nabla f_{ij}(\widetilde{\mathbf{x}}_{ij}^k)}{mp_{ij}} + \frac{\sum_{j=1}^{m}\nabla f_{ij}(\widetilde{\mathbf{x}}_{ij}^k) - \nabla f_{ij}(\mathbf{x}^*)}{m}\right\|^2$$

$$= \|\mathbf{X}^k - \mathbf{X}^*\|^2 - 2\eta\sum_{i=1}^{n}\langle\mathbf{x}_i^k - \mathbf{x}^*, \nabla f_i(\mathbf{x}_i^k)\rangle + 2\eta\sum_{i=1}^{n}(f_i(\mathbf{x}_i^k) - f_i(\mathbf{x}^*) - \mathrm{V}_{f_i}(\mathbf{x}_i^k, \mathbf{x}^*))$$

$$+ \eta^2\sum_{i=1}^{n}\sum_{j=1}^{m}p_{ij}\left\|\frac{\nabla f_{ij}(\mathbf{x}_i^k) - \nabla f_{ij}(\mathbf{x}^*) + f_{ij}(\mathbf{x}^*) - \nabla f_{ij}(\widetilde{\mathbf{x}}_{ij}^k)}{mp_{ij}} + \frac{\sum_{j=1}^{m}\nabla f_{ij}(\widetilde{\mathbf{x}}_{ij}^k) - \nabla f_{ij}(\mathbf{x}^*)}{m}\right\|^2.$$

Using the strong convexity of $f_{ij}$ in Assumption 4 and applying $(a+b)^2 \leq 2a^2 + 2b^2$ to the last term, we have

$$\mathbb{E}\|\mathbf{X}^k - \mathbf{X}^* - \eta\mathbf{G}^k + \eta\nabla\mathbf{F}(\mathbf{X}^*)\|^2$$
$$\leq (1 - \mu\eta)\|\mathbf{X}^k - \mathbf{X}^*\| - 2\eta\sum_{i=1}^{n} \mathbf{V}_{f_i}(\mathbf{x}_i^k, \mathbf{x}^*) + 2\eta^2\sum_{i=1}^{n}\sum_{j=1}^{m}\frac{1}{m^2 p_{ij}}\|\nabla f_{ij}(\mathbf{x}_i^k) - \nabla f_{ij}(\mathbf{x}^*)\|^2$$
$$+ 2\eta^2\sum_{i=1}^{n}\sum_{j=1}^{m} p_{ij}\left\|\frac{\nabla f_{ij}(\mathbf{x}^*) - \nabla f_{ij}(\tilde{\mathbf{x}}_{ij}^k)}{m p_{ij}} + \frac{\sum_{j=1}^{m}\nabla f_{ij}(\tilde{\mathbf{x}}_{ij}^k) - \nabla f_{ij}(\mathbf{x}^*)}{m}\right\|^2.$$

Let $u_i$ be the random variable taking values in $\left\{\frac{1}{m p_{il}}(\nabla f_{il}(\tilde{\mathbf{x}}_{il}^k) - \nabla f_{il}(\tilde{\mathbf{x}}^*)) : l \in [m]\right\}$ with distribution $\mathcal{P}_i = \{p_{il} : l \in [m]\}$, then the last term of the above inequality can be upper bounded by

$$2\eta^2\sum_{i=1}^{n}\mathbb{E}\|u_i - \mathbb{E}u_i\|^2 \leq 2\eta^2\sum_{i=1}^{n}\mathbb{E}\|u_i\|^2$$
$$= 2\eta^2\sum_{i=1}^{n}\sum_{j=1}^{m} p_{ij}\left\|\frac{1}{m p_{ij}}(\nabla f_{ij}(\tilde{\mathbf{x}}_{ij}^k) - \nabla f_{ij}(\tilde{\mathbf{x}}^*))\right\|^2.$$

Combining the inequality with the upper bound, we get

$$\mathbb{E}\|\mathbf{X}^k - \mathbf{X}^* - \eta\mathbf{G}^k + \eta\nabla\mathbf{F}(\mathbf{X}^*)\|^2$$
$$\leq (1 - \mu\eta)\|\mathbf{X}^k - \mathbf{X}^*\| - 2\eta\sum_{i=1}^{n} \mathbf{V}_{f_i}(\mathbf{x}_i^k, \mathbf{x}^*) + \frac{4\eta^2 L}{m^2}\sum_{i=1}^{n}\sum_{j=1}^{m}\frac{1}{p_{ij}}\mathbf{V}_{f_{ij}}(\mathbf{x}_i^k, \mathbf{x}^*)$$
$$+ 2\eta^2\sum_{i=1}^{n}\sum_{j=1}^{m}\frac{1}{m^2 p_{ij}}\|\nabla f_{ij}(\tilde{\mathbf{x}}_{ij}^k) - \nabla f_{ij}(\mathbf{x}^*)\|^2$$
$$\leq (1 - \mu\eta)\|\mathbf{X}^k - \mathbf{X}^*\| - 2\eta\sum_{i=1}^{n} \mathbf{V}_{f_i}(\mathbf{x}_i^k, \mathbf{x}^*) + \frac{4\eta^2 L}{m^2}\sum_{i=1}^{n}\sum_{j=1}^{m}\frac{1}{p_{ij}}\mathbf{V}_{f_{ij}}(\mathbf{x}_i^k, \mathbf{x}^*)$$
$$+ \frac{4\eta^2 L}{m^2}\sum_{i=1}^{n}\sum_{j=1}^{m}\frac{1}{p_{ij}}\mathbf{V}_{f_{ij}}(\tilde{\mathbf{x}}_{ij}^k, \mathbf{x}^*),$$

where the first inequality uses the Lipschitz smoothness of $f_{ij}$ in Assumption 4.

From the update of $\tilde{\mathbf{x}}_{ij}$, we have

$$\mathbb{E}\mathbf{V}_{f_{ij}}(\tilde{\mathbf{x}}_{ij}^{k+1}, \mathbf{x}^*) = p_{ij}\mathbf{V}_{f_{ij}}(\mathbf{x}_i^k, \mathbf{x}^*) + (1 - p_{ij})\mathbf{V}_{f_{ij}}(\tilde{\mathbf{x}}_{ij}^k, \mathbf{x}^*).$$

Combine it with the key inequality of Lemma 4,

$$\mathbb{E}\Phi^{k+1} + \tilde{c}\sum_{i=1}^{n}\sum_{j=1}^{m}\mathbb{E}V_{f_{ij}}(\tilde{\mathbf{x}}_{ij}^{k+1}, \mathbf{x}^*)$$

$$\leq (1-\eta\mu)\|\mathbf{X}^k - \mathbf{X}^*\|^2 - 2\eta\sum_{i=1}^{n}\sum_{j=1}^{m}\frac{1}{m}V_{f_{ij}}(\mathbf{x}_i^k, \mathbf{x}^*) + \frac{4\eta^2 L}{m^2}\sum_{i=1}^{n}\sum_{j=1}^{m}\frac{1}{p_{ij}}V_{f_{ij}}(\mathbf{x}_i^k, \mathbf{x}^*)$$

$$+ \tilde{c}\sum_{i=1}^{n}\sum_{j=1}^{m}p_{ij}V_{f_{ij}}(\mathbf{x}_i^k, \mathbf{x}^*) + \tilde{c}\sum_{i=1}^{n}\sum_{j=1}^{m}(1-p_{ij})V_{f_{ij}}(\tilde{\mathbf{x}}_{ij}^k, \mathbf{x}^*) + \frac{4\eta^2 L}{m^2}\sum_{i=1}^{n}\sum_{j=1}^{m}\frac{1}{p_{ij}}V_{f_{ij}}(\tilde{\mathbf{x}}_{ij}^k, \mathbf{x}^*)$$

$$+ \frac{2\eta^2}{\gamma}\left(1 - \frac{\gamma}{2}\lambda_{\min}(\mathbf{I} - \mathbf{W})\right)\|\mathbf{D}^k - \mathbf{D}^*\|^2_{(\mathbf{I}-\mathbf{W})^\dagger} + \sqrt{C}(1-\alpha)\|\mathbf{H}^k - \mathbf{Z}^*\|^2$$

$$\leq (1-\eta\mu)\|\mathbf{X}^k - \mathbf{X}^*\|^2 + \tilde{c}\sum_{i=1}^{n}\sum_{j=1}^{m}(1-p_{ij})V_{f_{ij}}(\tilde{\mathbf{x}}_{ij}^k, \mathbf{x}^*) + \frac{4\eta^2 L}{m^2}\sum_{i=1}^{n}\sum_{j=1}^{m}\frac{1}{p_{ij}}V_{f_{ij}}(\tilde{\mathbf{x}}_{ij}^k, \mathbf{x}^*)$$

$$+ \frac{2\eta^2}{\gamma}\left(1 - \frac{\gamma}{2}\lambda_{\min}(\mathbf{I} - \mathbf{W})\right)\|\mathbf{D}^k - \mathbf{D}^*\|^2_{(\mathbf{I}-\mathbf{W})^\dagger} + \sqrt{C}(1-\alpha)\|\mathbf{H}^k - \mathbf{Z}^*\|^2,$$

where the last inequality is guaranteed by

$$\frac{2\eta}{m} - \frac{4\eta^2 L}{m^2 p_{ij}} - \tilde{c}p_{ij} \geq 0.$$

Define $p_{\min} = \min_{i,j}\{p_{ij}\}$ and take $\tilde{c} = \frac{8\eta^2 L}{m^2 p_{\min}^2}$, then by the choice of $\{p_{ij}\}$ and $\eta$, the above condition is satisfied due to

$$\frac{2\eta}{m} - \frac{4\eta^2 L}{m^2 p_{ij}} - \tilde{c}p_{ij} = \frac{1}{3Lm} - \frac{1}{9Lm} - \frac{2}{9Lm} = 0.$$

Note that

$$\tilde{c}(1-p_{ij}) + \frac{4\eta^2 L}{m^2 p_{ij}} = \tilde{c}\left(1 - p_{ij} + \frac{\frac{4\eta^2 L}{m^2 p_{ij}}}{\tilde{c}}\right) = \tilde{c}\left(1 - \frac{1}{2m}\right).$$

Therefore,

$$\mathbb{E}\Phi^{k+1} + \frac{2}{9L}\sum_{i=1}^{n}\sum_{j=1}^{m}V_{f_{ij}}(\tilde{\mathbf{x}}_{ij}^{k+1}, \mathbf{x}^*)$$

$$\leq \max\left\{\frac{1-\eta\mu}{M}, 1 - \frac{\gamma}{2}\lambda_{\min}(\mathbf{I} - \mathbf{W}), 1 - \alpha, 1 - \frac{1}{2m}\right\}\left(\Phi^k + \frac{2}{9L}\sum_{i=1}^{n}\sum_{j=1}^{m}V_{f_{ij}}(\tilde{\mathbf{x}}_{ij}^k, \mathbf{x}^*)\right).$$

**Complexity** The complexity analysis is identical to that in Theorem 8 with the single exception that we replace $p$ by $m^{-1}$ so we omit it here. ∎

Using $\widetilde{\mathcal{O}}(\cdot)$ as the abbreviation of $\mathcal{O}((\cdot)\log(1/\epsilon))$, we simplify the above two theorems as the following corollary.

**Corollary 10** *We can achieve $\|\mathbf{x}_i^k - \mathbf{x}^*\|^2 \le \epsilon$ in expectation on each node after the number of iterations*

- $$K = \widetilde{\mathcal{O}}\left(\sqrt{C}(1+C)\kappa_f\kappa_g + (1+C)(\kappa_f + \kappa_g) + p^{-1}\right)$$

  *for Loopless SVRG and*

- $$K = \widetilde{\mathcal{O}}\left(\sqrt{C}(1+C)\kappa_f\kappa_g + (1+C)(\kappa_f + \kappa_g) + m\right)$$

  *for SAGA.*

**Remark 11** *In particular, when there is no compression, i.e., $C = 0$ and the network is fully connected, i.e., $\kappa_g = 1$, the complexity of Loopless SVRG is reduced to $\widetilde{\mathcal{O}}\left(\kappa_f + p^{-1}\right)$, which matches that given in Kovalev et al. (2020) and the complexity of SAGA is reduced to $\widetilde{\mathcal{O}}\left(\kappa_f + m\right)$ shown in Defazio et al. (2014).*

### 4.3 Connection with Existing Algorithms

In Section 2, we have discussed the motivation of LEAD and Prox-LEAD. We now turn to the relation between LEAD and some other existing algorithms from the perspective of the dual problem. We look at the problem (2) without the non-smooth regularizer, i.e., $\mathbf{R}(\mathbf{X}) = \mathbf{0}$. Consider the Fenchel conjugate of $\mathbf{F}$ defined as $\mathbf{F}^*(\mathbf{Y}) = \sup_{\mathbf{X} \in \mathbb{R}^{n \times p}} \mathbf{Y}^\top \mathbf{X} - \mathbf{F}(\mathbf{X})$, then we obtain the dual problem as

$$\min_{\mathbf{S} \in \mathbb{R}^{n \times p}} \mathbf{F}^*(-\sqrt{\mathbf{I} - \mathbf{W}}\mathbf{S}).$$

If we apply the gradient descent method to the above problem, we need to evaluate the gradient $-\sqrt{\mathbf{I} - \mathbf{W}}\nabla\mathbf{F}^*(-\sqrt{\mathbf{I} - \mathbf{W}}\mathbf{S})$. Let $\mathbf{D} = \sqrt{\mathbf{I} - \mathbf{W}}\mathbf{S}$, then the iteration follows

$$\mathbf{D}^{k+1} = \mathbf{D}^k + \theta(\mathbf{I} - \mathbf{W})\nabla\mathbf{F}^*(-\mathbf{D}^k).$$

When the gradient of the dual function is available, the communication proceeds after the gradient evaluation. If we compress the only communication step, the algorithm leads to Option A in Kovalev et al. (2021a) with the single exception that the quantization procedure is slightly different. Since $\mathbf{D}$ belongs to the dual space of the original optimized variable, we derive the solution of the primal problem via the relation

$$\mathbf{X}^{k+1} = \nabla\mathbf{F}^*(-\mathbf{D}^k) \tag{20}$$

and the linear convergence is guaranteed under the strongly convex assumption on $\mathbf{F}$.

In most cases, it is difficult to evaluate the conjugate function, so we rewrite the relation (20) into the following minimization problem

$$\mathbf{X}^{k+1} = \operatorname*{arg\,min}_{\mathbf{X} \in \mathbb{R}^{n \times p}} \mathbf{F}(\mathbf{X}) + \langle \mathbf{D}^k, \mathbf{X} \rangle,$$

and get an inexact estimate using the gradient method. Applying one step of gradient descent method to the subproblem and inserting the update into the original dual iterations, we get the following primal-dual iteration

$$
\left\lfloor
\begin{array}{l}
\mathbf{X}^{k+1} = \mathbf{X}^k - \eta \nabla \mathbf{F}(\mathbf{X}^k) - \eta \mathbf{D}^k, \\
\mathbf{D}^{k+1} = \mathbf{D}^k + \theta (\mathbf{I} - \mathbf{W}) \mathbf{X}^{k+1},
\end{array}
\right.
$$

where $\eta, \theta$ are stepsizes for primal and dual update respectively. It can be shown that this iteration is a special case of the incremental primal-dual gradient method (PDGM) in Alghunaim and Sayed (2020) and the linear convergence rate can be guaranteed. Furthermore, when the communication of $\mathbf{X}^{k+1}$ in the dual update is conducted by the compression procedure (and the stochastic gradient estimation is involved), the algorithm recovers Option B (Option C) of LessBit in Kovalev et al. (2021a).

If we proceed one more gradient descent step in the subproblem, we get

$$
\left\lfloor
\begin{array}{l}
\mathbf{X}^{k+1} = \mathbf{X}^k - \eta \nabla \mathbf{F}(\mathbf{X}^k) - \eta \mathbf{D}^k, \\
\overline{\mathbf{X}}^{k+1} = \mathbf{X}^{k+1} - \eta \nabla \mathbf{F}(\mathbf{X}^{k+1}) - \eta \mathbf{D}^k, \\
\mathbf{D}^{k+1} = \mathbf{D}^k + \theta (\mathbf{I} - \mathbf{W}) \overline{\mathbf{X}}^{k+1}.
\end{array}
\right.
$$

The addition of the step does not increase the computation of the gradient $\nabla \mathbf{F}$ because it can be reused in the next iteration. So, we switch the order of the iteration and derive

$$
\left\lfloor
\begin{array}{l}
\overline{\mathbf{X}}^{k+1} = \mathbf{X}^k - \eta \nabla \mathbf{F}(\mathbf{X}^k) - \eta \mathbf{D}^k, \\
\mathbf{D}^{k+1} = \mathbf{D}^k + \theta (\mathbf{I} - \mathbf{W}) \overline{\mathbf{X}}^{k+1}, \\
\mathbf{X}^{k+1} = \mathbf{X}^k - \eta \nabla \mathbf{F}(\mathbf{X}^k) - \eta \mathbf{D}^{k+1}.
\end{array}
\right.
$$

By setting $\theta = 1$, the above algorithm recovers NIDS of the smooth problems in Li et al. (2019); Li and Yan (2021). It has been shown that the additional step in NIDS improves the linear convergence rate of the previous two algorithms in terms of the condition numbers of the objective function and the network. The detailed comparison is listed in Table 3.

| Algorithm | R | $\nabla$F | Comp. | Convergence complexity |
|---|---|---|---|---|
| Dual Gradient Descent | ✗ | ✗ | ✗ | $\widetilde{\mathcal{O}}(\kappa_f \kappa_g)$ |
| LessBit-Option A<br>Kovalev et al. (2021a) | ✗ | ✗ | ✓ | $\widetilde{\mathcal{O}}(C + \kappa_f \kappa_g + C \kappa_f \widetilde{\kappa_g})$ |
| PDGM<br>Alghunaim and Sayed (2020) | ✗ | ✓ | ✗ | $\widetilde{\mathcal{O}}(\kappa_f + \kappa_f \kappa_g)$ |
| LessBit-Option B<br>Kovalev et al. (2021a) | ✗ | ✓ | ✓ | $\widetilde{\mathcal{O}}(C + \kappa_f \kappa_g + C \kappa_f \widetilde{\kappa_g})$ |
| NIDS<br>Li and Yan (2021) | ✗ | ✓ | ✗ | $\widetilde{\mathcal{O}}(\kappa_f + \kappa_g)$ |
| LEAD<br>Liu et al. (2021) | ✗ | ✓ | ✓ | $\widetilde{\mathcal{O}}((1 + C)(\kappa_f + \kappa_g) + C \kappa_f \kappa_g)$ |
| PUDA<br>Alghunaim et al. (2020) | ✓ | ✓ | ✗ | $\widetilde{\mathcal{O}}(\kappa_f + \kappa_g)$ |
| **Prox-LEAD**<br>**this paper, Algorithm 1** | ✓ | ✓ | ✓ | $\widetilde{\mathcal{O}}((1 + C)(\kappa_f + \kappa_g) + \sqrt{C}(1 + C)\kappa_f \kappa_g)$ |

Table 3: The comparison of algorithms mentioned in Section 4.3; $\kappa_f := \frac{L}{\mu}, \kappa_g := \frac{\lambda_{\max}(\mathbf{I} - \mathbf{W})}{\lambda_{\min}(\mathbf{I} - \mathbf{W})}$ and $\widetilde{\kappa_g} := \frac{\max_{(i,j) \in \mathcal{E}} 1 - w_{ij}}{\lambda_{\min}(\mathbf{I} - \mathbf{W})}$. Comp. stands for the compressed communication.

## 5. Numerical Experiments

In this section, we present numerical experiments to validate the convergence of the proposed algorithms, including LEAD in the smooth case and Prox-LEAD in the non-smooth case, as well as their stochastic variants.

### 5.1 Experimental Setting

**Baselines.** To demonstrate the effectiveness of the proposed algorithms, we compare them with the following baselines: 1) two state-of-the-art decentralized algorithms with compression: Choco (Koloskova et al., 2019) and LessBit (Kovalev et al., 2021b); 2) three non-compressed algorithms: DGD (Yuan et al., 2016), NIDS (Li et al., 2019), and P2D2 (Alghunaim et al., 2019). Note that NIDS and P2D2 support the non-smooth case. In the stochastic case, we also include LessBit with Option C (LessBit-SGD) and Option D (LessBit-LSVRG).

**Setup.** We consider eight machines connected in a ring topology network. Each agent can only exchange information with its two 1-hop neighbors. The mixing weight is simply set as $1/3$. For compression, we use the unbiased $b$-bits quantization method with $\infty$-norm

$$Q_\infty(\mathbf{x}) := \left(\|\mathbf{x}\|_\infty 2^{-(b-1)} \text{sign}(\mathbf{x})\right) \cdot \left\lfloor \frac{2^{(b-1)}|\mathbf{x}|}{\|\mathbf{x}\|_\infty} + \mathbf{u} \right\rfloor, \tag{21}$$

where $\cdot$ is the Hadamard product, $|\mathbf{x}|$ is the elementwise absolute value of $\mathbf{x}$, and $\mathbf{u}$ is a random vector uniformly distributed in $[0, 1]^p$. Only $\text{sign}(\mathbf{x})$, norm $\|\mathbf{x}\|_\infty$, and integers in the bracket need to be transmitted. Note that this quantization method is similar to the quantization used in QSGD (Alistarh et al., 2017) and Choco (Koloskova et al., 2019), but we use the $\infty$-norm scaling instead of the 2-norm. This small change brings significant improvement on compression precision as justified both theoretically and empirically in Appendix C in (Liu et al., 2021). In this section, we choose 2-bit quantization and quantize the data in a blockwise manner (block size = 256).

For all experiments, we tune the stepsize $\eta$ in the range $[0.01, 0.1]$. For LEAD and Prox-LEAD, we simply fix $\alpha = 0.5$ and $\gamma = 1.0$ for all experiments since they are very robust to the parameter settings. The parameters $\gamma$ in Choco and $\theta$ in LessBit are tuned from $\{0.01, 0.05, 0.1, 0.2, 0.5, 0.8, 1.0\}$.

**Logistic regression.** We consider a regularized logistic regression problem with a cross-entropy objective function:

$$f(\mathbf{X}) = -\frac{1}{m} \sum_{i=1}^{m} \sum_{j=1}^{C} \mathbf{y}_{i,j} \log(\mathbf{a}_i^\top \mathbf{X}_j) + \lambda_1 \|\mathbf{X}\|_1 + \lambda_2 \|\mathbf{X}\|_2^2,$$

where $\mathbf{a}_i \in \mathbb{R}^p, \mathbf{y} \in \mathbb{R}^{m \times C}, \mathbf{X} \in \mathbb{R}^{p \times C}$ and $C$ is the number of classes. We use the MNIST dataset and distribute the samples equally to all the machines in a non-iid way, sorted by their labels. Note that this is the heterogeneous data settings where the data distribution from each agent is very different, which is more challenging than the homogeneous data setting where all the agents share the same data distribution. In the smooth case, we set $\lambda_1 = 0$ and $\lambda_2 = 0.005$, and in the non-smooth case, we set $\lambda_1 = 0.005$ and $\lambda_2 = 0.005$. In the stochastic case, the training data in each agent are evenly divided into 15 mini-batches. The performance is measured by the the training suboptimality, i.e., $\|\mathbf{X}^k - \mathbf{X}^*\|_F^2$, where $\mathbf{X}^*$ denotes the optimal solution.

**Smooth case.** The experiments in the smooth case are showed in Fig. 1. From Fig. 1a, we can observe that when full gradients are available, NIDS, LessBit and LEAD converge linearly to the optimal solution, while DGD and Choco suffer from the convergence bias. Fig. 1b demonstrates the benefit of communication compression when considering the suboptimality with respect to the communication bits. Note that the performance of LEAD (32bit) matches well with LEAD (2bit), which validates that the compression doesn't hurt the convergence for LEAD, while the communication bits are significantly reduced.

Fig. 1c and Fig. 1d show the performance with stochastic gradients. We can make the following observations: 1) The performance of LEAD-SGD (2bit), LEAD-SAGA (2bit) and LEAD-LSVRG (2bit) match well with LEAD-SGD (32bit), LEAD-SAGA (32bit) and LEAD-LSVRG (32bit), respectively, which indicates that the compression in LEAD doesn't hurt the convergence; 2) The linear convergence of the variance-reduction variants, such as LEAD-SAGA (2bit) and LEAD-LSVRG (2bit), verifies our theoretical analyses[2]; 3) LEAD-SAGA (2bit) and LEAD-LSVRG (2bit) significantly outperform all baselines [3]. The benefit of communication compression can be clearly illustrated by Fig. 1d.

---

[2]LEAD-SAGA outperform LEAD-LSVRG in terms of the number of gradient computation since LEAD-SAGA computes fewer gradient evaluations in each iteration by sacrificing the memory space. However, LEAD-LSVRG outperforms LEAD-SAGA in terms of communication bits since the extra gradient computation in LEAD-LSVRG doesn't increase communication cost but it improves the convergence speed. Similar phenomenon is also observed in the non-smooth case in Figure 2c and Fig. 2d.

[3]LEAD-SGD (2bit) and LEAD-LSVRG (2bit) outperform LessBit-SGD (2bit) and LessBit-LSVRG (2bit), which shows the advantages of the extra gradient descent step in LEAD, as discussed in Section 4.3.

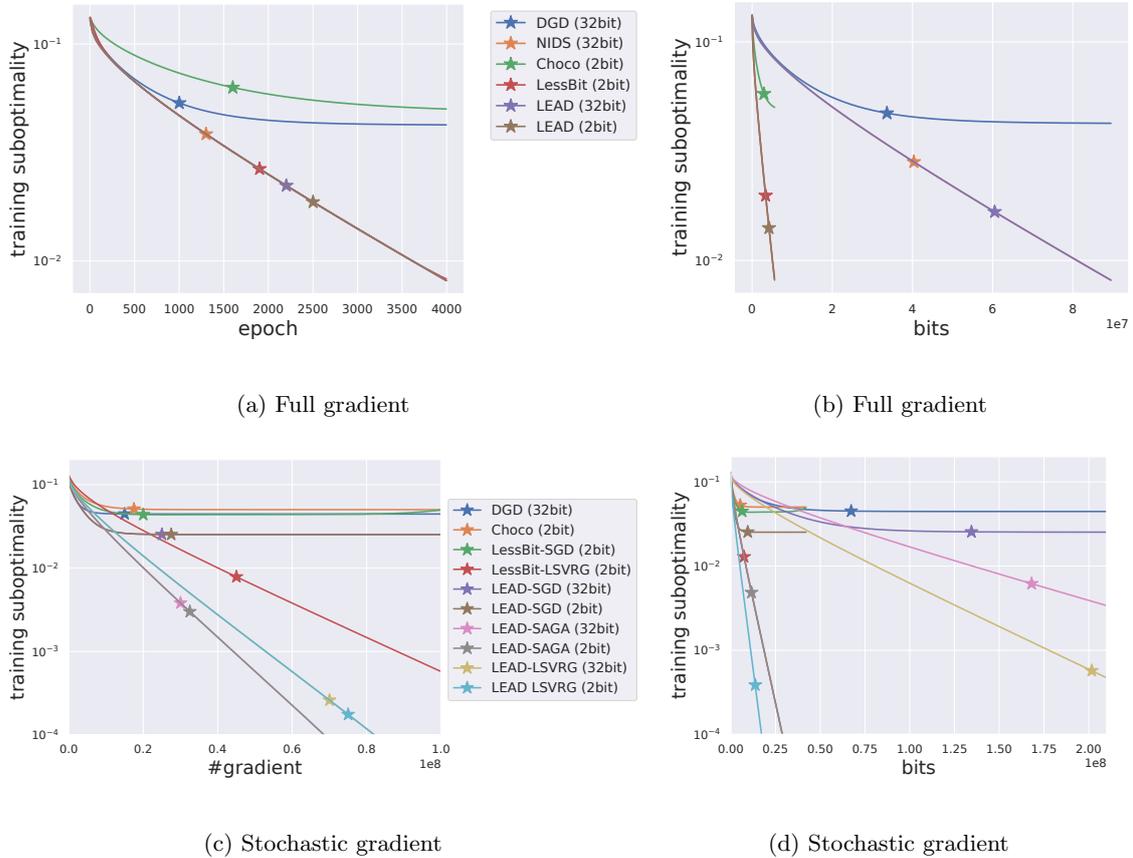

(a) Full gradient

(b) Full gradient

(c) Stochastic gradient

(d) Stochastic gradient

Figure 1: Smooth logistic regression problem ($\lambda_1 = 0$). In the full gradient case ((a) and (b)), LEAD (2bit) and LessBit (2bit) converge similarly as NIDS (32bit) and LEAD (32bit) in terms of epochs/iterations, but they requires much fewer bits in communication. In the stochastic case ((c) and (d)), the 2bit variants of LEAD match well with their 32bit variants in terms of the number of gradient evaluation, but they requires much fewer bits. Note that LEAD-SAGA requires more memory and more iterations/communication than LEAD-LSVRG, but it computes only one gradient in each iteration, while LEAD-LSVRG computes at least two gradients in each iteration.

**Non-smooth case.** The experiments in the non-smooth case are showed in Fig. 2. Fig. 2a shows that Prox-LEAD (2bit) achieves linear convergence to the optimal solution with full gradient, and its performance matches well with the non-compressed version Prox-LEAD (32bit). It also converges similarly with other non-compressed baselines such as P2D2 and NIDS. Fig. 2b demonstrates the tremendous advantages of communication compression in Prox-LEAD (2bit) when considering the communication bits.

---

Though LessBit-LSVRG (2bit) has the same communication bit as LEAD-SAGA (2bit), LEAD-SAGA (2bit) requires about half of the gradient evaluation as LessBit-LSVRG (2bit).

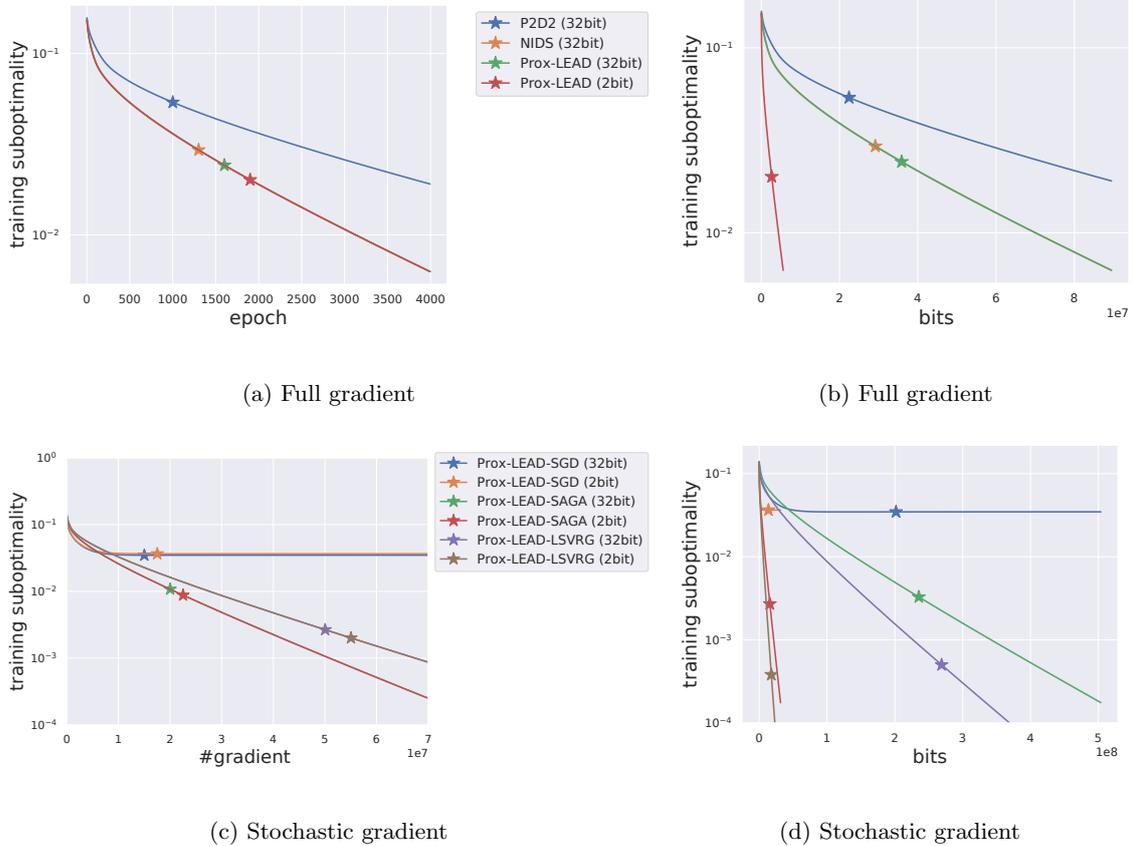

(a) Full gradient

(b) Full gradient

(c) Stochastic gradient

(d) Stochastic gradient

Figure 2: Non-smooth logistic regression problem ($\lambda_1 = 0.005$). In the full gradient case ((a) and (b)), Prox-LEAD (2bit) converges similarly as NIDS and Prox-LEAD (32bit) in terms of epochs/iterations, but it requires much fewer bits than the other three algorithms. In the stochastic case ((c) and (d)), the 2bit variants match well with their 32bit variants in terms of the number of gradient evaluation, but they requires much fewer bits. Note that Prox-LEAD-SAGA requires more memory and more iterations/communication than Prox-LEAD-LSVRG, but it computes only one gradient in each iteration, while Prox-LEAD-LSVRG computes at least two gradients in each iteration.

Fig. 2c and Fig. 2d present the performance with stochastic gradients. It can be observed from Fig. 2c that: 1) Prox-LEAD-SAGA (2bit) and Prox-LEAD-LSVRG (2bit) maintains linear convergence with communication compression and stochastic gradients; 2) The compressed versions of Prox-LEAD all match well with the non-compressed versions. Fig. 2b shows that the advantages of communication compression in Prox-LEAD are very significant in terms of communication bits.

To summarize, the experiments in this section verify the theoretical linear convergence of the proposed algorithm when the nonsmooth objective, stochastic gradients and commu-nication compression are present. They also suggest the state-of-the-art performance in the comparison with strong baseline algorithms.

## 6. Conclusion

In this paper, we consider the decentralized stochastic composite optimization problem. A decentralized proximal stochastic gradient algorithm with communication compression, Prox-LEAD, is proposed to improve the communication efficiency and convergence rates. We provide rigorous theoretical analyses and convergence complexities for the proposed algorithm in the general stochastic setting and the finite-sum setting. We establish the linear convergence rate with variance reduction schemes and well-controlled compression error. Both the theorems and numerical experiments demonstrate the effectiveness of Prox-LEAD in reducing the communication cost and the advantages over existing algorithms. Moreover, our algorithmic framework builds bridges between many known algorithms, and it potentially enlightens the communication compression for other primal-dual algorithms.

## Acknowledgments

Xiaorui Liu and Dr. Jiliang Tang are supported by the National Science Foundation (NSF) under grant numbers CNS-1815636, IIS-1928278, IIS-1714741, IIS-1845081, IIS-1907704, IIS-1955285, and the Army Research Office (ARO) under grant number W911NF-21-1-0198. Yao Li and Dr. Ming Yan are supported by NSF grant DMS-2012439 and Facebook Faculty Research Award (Systems for ML).